\documentclass[letterpaper, 10 pt, conference]{ieeeconf}  
\IEEEoverridecommandlockouts  
\usepackage{graphicx} 
\usepackage{amsmath} 
\usepackage{amssymb}  
\usepackage{times}

\usepackage{array}
\usepackage{url} 
\usepackage{enumerate}
\usepackage{multirow} 
\usepackage{float}
\usepackage{paralist}
\usepackage{amsfonts}
\usepackage{subcaption}
\usepackage{caption}
\usepackage{longtable}
\usepackage{algorithm} 
\usepackage{algpseudocode}
\usepackage{wrapfig}
\usepackage{mathtools, cuted}

\newcolumntype{L}[1]{>{\raggedright\let\newline\\\arraybackslash\hspace{0pt}}m{#1}}
\newcolumntype{C}[1]{>{\centering\let\newline\\\arraybackslash\hspace{0pt}}m{#1}}
\newcolumntype{R}[1]{>{\raggedleft\let\newline\\\arraybackslash\hspace{0pt}}m{#1}}

\DeclareMathOperator*{\argmax}{arg\,max}

\newcommand{\nostarnote}[1]{}
\newcommand{\nonote}[1]{}

\title{WPE - Understanding Human Motion} 
\title{\LARGE \bf Understanding Human Motion and Gestures for Underwater Human-Robot Collaboration$^{*}$}%
\author{
    Md Jahidul Islam \\ \small Interactive Robotics and Vision Laboratory,\\Department of Computer Science and Engineering,\\University of Minnesota- Twin Cities, US. \\ {\tt\footnotesize islam034@umn.edu}%
    \thanks{$^{*}$This report is based on the published papers~\cite{islam2018dynamic} and~\cite{islam2017mixed}. Md Jahidul Islam is the primary investigator of the report, he is supervised by Junaed Sattar, Assistant Professor, Department of Computer Science and Engineering, University of Minnesota Twin Cities.%
    }
}

\begin{document}

\maketitle
\thispagestyle{empty}
\pagestyle{empty}

\begin{abstract}
In this paper, we present a number of robust methodologies for an underwater robot to visually detect, follow, and interact with a diver for collaborative task execution. We design and develop two autonomous diver-following algorithms, the first of which utilizes both spatial- and frequency-domain features pertaining to human swimming patterns in order to visually track a diver. The second algorithm uses a convolutional neural network-based model for robust tracking-by-detection. In addition, we propose a hand gesture-based human-robot communication framework that is syntactically simpler and computationally more efficient than the existing grammar-based frameworks. In the proposed interaction framework, deep visual detectors are used to provide accurate hand gesture recognition; subsequently, a finite-state machine performs robust and efficient gesture-to-instruction mapping. The distinguishing feature of this framework is that it can be easily adopted by divers for communicating with underwater robots without using artificial markers or requiring memorization of complex language rules. Furthermore, we validate the performance and effectiveness of the proposed methodologies through extensive field experiments in closed- and open-water environments. Finally, we perform a user interaction study to demonstrate the usability benefits of our proposed interaction framework compared to existing methods.
\end{abstract}

\section{INTRODUCTION}\label{sec:intro}
Underwater robotics is an area of increasing importance, with existing and emerging applications ranging from inspection and surveillance, to data collection and mapping tasks. Since truly autonomous underwater navigation is still an open problem, underwater missions often require a team of human divers and autonomous robots to cooperatively perform tasks. The human divers typically lead the missions and operate the robots during mission execution~\cite{islam2018person}. Such situations arise in numerous important applications such as undersea pipeline and ship-wreck inspection, marine life and seabed monitoring, and many other exploration activities~\cite{sattar2008enabling}.

\begin{figure}[ht]
    \vspace{1mm}
    \centering
    \begin{subfigure}[t]{0.22 \textwidth}
        \includegraphics[width=\linewidth]{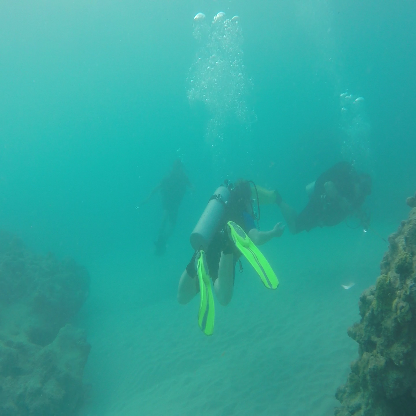} 
        \caption{While following a diver}
    \end{subfigure}
    \begin{subfigure}[t]{0.25\textwidth}
        \includegraphics[width=\linewidth]{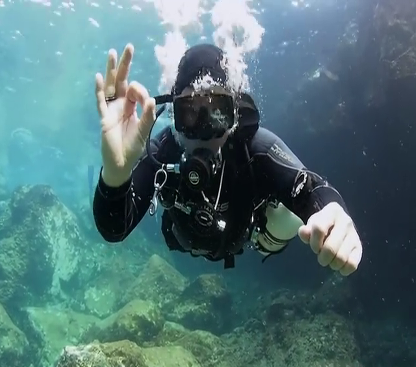}
        \caption{Getting hand gesture-based instructions from a diver}
    \end{subfigure}
    \vspace{-1mm}
    \caption{Views from the camera of an underwater robot during a cooperative reef exploration task.}
 \vspace{-3mm}
    \label{fig:intro}
    \end{figure}
    
Without sacrificing the generality of the applications, we consider a single-robot setting where a human diver leads and interacts with the robot at certain stages of an underwater mission. The robot follows the diver and performs the tasks instructed by the diver during the operation. Such semi-autonomous behavior of a mobile robot with human-in-the-loop guidance reduces operational overhead by eliminating the necessity of teleoperation~\cite{sattar2008enabling}. In addition, the ability to dynamically guide the robot and reconfigure its program parameters is important for underwater exploration and data collection processes. However, since Wi-Fi or radio (\textit{i.e.}, electromagnetic) communication is severely degraded underwater~\cite{dudek2008sensor}, the current task needs to be interrupted, and the robot needs to be brought to the surface in order to reconfigure the mission parameters. This is inconvenient and often expensive in terms of time and physical resources. Therefore, triggering parameter changes based on human input while the robot is underwater, without requiring a trip to the surface, is a simpler and more efficient alternative approach. 

Visual perception is challenging in underwater environments due to the unfavorable visual conditions arising from generally degraded optics caused by factors such as limited visibility, variations in illumination, chromatic distortions, etc. A practical alternative is to use acoustic sensors such as sonars and hydrophones. However, their applicability and feasibility in interactive applications are limited; hence, they are only used for tracking applications \cite{mandic2016underwater, demarco2013sonar}. Additionally, acoustic sensors face challenges in coastal waters due to scattering and reverberation. Furthermore, their use is often limited by government regulations on the sound level in marine environments~\cite{islam2018person}. These are compelling reasons why visual sensing is more feasible and generally applicable for underwater applications.

In this paper, we focus on enabling the computational capabilities of an underwater robot for it to operate in human-robot cooperative settings using visual sensing (\textit{e.g.}, Figure \ref{fig:intro}). In particular, we develop methodologies for understanding human swimming motion and hand gesture-based instructions. Specifically, we make the following contributions in this paper: 
\begin{itemize}
\item We design robust and efficient algorithms for an underwater robot to autonomously follow a diver. 
\item Additionally, we propose a hand gesture-based human-robot communication framework that is syntactically simpler and computationally more efficient than existing grammar-based frameworks.
\item We evaluate the proposed methodologies through extensive field experiments. The experiments are performed in closed-water and open-water environments on an underwater robot. 
\end{itemize}

We first consider the autonomous diver-following problem, where the robot needs to visually detect and follow a diver swimming in an arbitrary direction. We develop two diver-following methodologies. The first method detects the motion directions of a diver by keeping track of his/her positions through the image sequences over time. In this method~\cite{islam2017mixed}, a Hidden Markov Model (HMM)-based approach prunes the search-space of all potential motion directions relying on image intensities in the spatial-domain. The diver's motion signature is subsequently detected in a sequence of non-overlapping image sub-windows exhibiting human swimming patterns. The pruning step ensures efficient computation by avoiding exponentially large search-spaces, whereas the frequency-domain detection allows us to detect the diver's position and motion direction accurately. The second method uses deep visual features to detect divers in the RGB image-space. In this method, a convolutional neural network (CNN)-based model is trained on a large dataset of hand-annotated images that are collected from various diver-following applications. The trained model is invariant to the scale and appearance of divers (\textit{e.g.}, the color of the suit/flippers, swimming directions, etc.) and robust to noise and image distortions~\cite{fabbri2018enhancing}.

We then develop a simple interaction framework where a diver can use a set of intuitive and meaningful hand gestures to program new instructions for the accompanying robot or reconfigure existing program parameters ``on-the-fly''~\cite{islam2018dynamic}. In the proposed framework, a CNN-based model is designed for hand gesture recognition; we also explore the state-of-the-art deep object detectors such as Faster RCNN~\cite{renNIPS15fasterrcnn} and Single Shot MultiBox Detector (SSD) ~\cite{liu2016ssd} to further improve the accuracy and robustness of hand gesture recognition. Once the hand gestures are recognized, a finite-state machine-based deterministic model efficiently performs the gesture-to-instruction mapping. These mapping rules are intuitively designed so that they can be easily interpreted and adopted by the divers. The major advantage of this design is that a diver can communicate with underwater robots in a natural way using their hands, without using artificial tags such as fiducial markers, complex electronic devices or requiring memorization of a potentially complex set of language rules. Additionally, it relieves the divers of the task of carrying a set of markers, which, if lost, put the mission in peril.

Furthermore, we demonstrate that both the diver-following and interaction modules can be used in real-time for practical applications. We evaluate the effectiveness of our proposed methodologies through extensive experimental evaluations. We perform field experiments both in open-water and closed-water (\textit{i.e.}, oceans and pools, respectively) environments on an underwater robot. We also perform a user interaction study to validate the usability of the proposed human-robot interaction framework compared to existing methods.

\section{RELATED WORK}
The underwater domain poses unique challenges for artificial (as well as natural) sensing, particularly more so for vision. Visual perception is often difficult for underwater robots because of light scattering, absorption and refraction, as well as the presence of suspended particulates. These phenomena affect poor visual conditions, variations in lighting, and chromatic distortions. For an underwater robot to have accurate visual sensing, robustness to noisy sensory data, accuracy, and fast running times are absolute necessities. In the following discussion, we present the existing visual perception methodologies for autonomous diver-following and various human-robot interaction frameworks.    

\subsection{Autonomous Diver Following}
Due to the operational simplicity and fast running times, simple feature-based trackers~\cite{sattar2006performance, sattar2005visual} are often practical choices for autonomous diver following. For instance, color-based tracking algorithms perform binary image thresholding based on the color of a diver's flippers or suit. The thresholded binary image is then refined to track the centroid of the target (diver) using algorithms such as mean-shift, particle filters, etc. Ensemble learning methods such as Adaptive Boosting (AdaBoost) has also been used for diver tracking~\cite{sattar2009robust}; AdaBoost learns a strong tracker from a large number of simple feature-based trackers. Such ensemble methods are proven to be computationally inexpensive yet highly accurate in practice. Optical flow-based methods can also be utilized to track diver's motion from one image frame to another. Optical flow is typically measured between two temporally ordered frames using the well-known Horn and Schunk formulation \cite{inoue1992robot} driven by brightness and smoothness assumptions on the image derivatives. Therefore, as long as the target motion is spatially and temporally smooth, optical flow vectors can be reliably used for detection. Several other feature-based tracking algorithms and machine learning techniques have been investigated for diver tracking and underwater object tracking in general. However, these methods are applicable mostly in favorable visual conditions (\textit{e.g.}, in clear visibility and favorable lighting conditions). 

\begin{figure}[h]
\vspace{1mm}
 \centering
    \includegraphics[width=\linewidth]{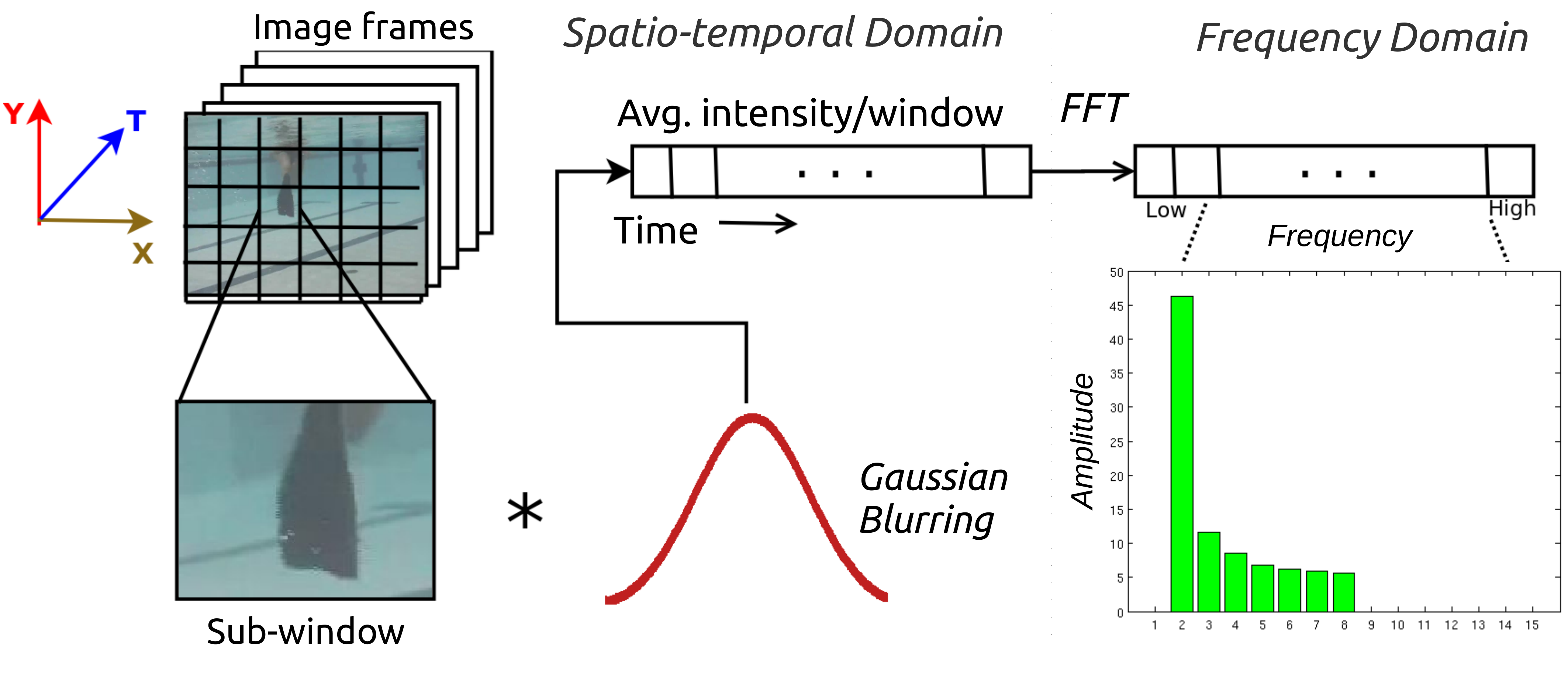}
 \caption{An outline of detecting periodic swimming signatures of a diver in the frequency-domain.}
 \vspace{-4mm}
 \label{fig:FourierT}
\end{figure}
    
Color distortions and low visibility issues are common in deep water scenarios. It is shown in~\cite{Sattar09RSS} that the human swimming cues in the frequency domain are stable and regular in noisy conditions. Specifically, intensity variations in the spatio-temporal domain along a diver's swimming direction have identifiable signatures in the frequency-domain. These intensity variations caused by a diver's swimming gait tend to generate high-energy responses in the $1$-$2$Hz frequency range. This inherent periodicity can be used as a cue for robust detection in noisy conditions; the overall process is outlined in Figure \ref{fig:FourierT}. The first contribution in this paper generalizes this idea in order to track arbitrary motions. Our proposed tracker uses spatial-domain features to keep track of a diver's potential motion directions using a Hidden Markov Model (HMM). Subsequently, it inspects the frequency-domain responses along those motion directions to find the most probable one to contain a diver's swimming trajectory. We name this algorithm the \textbf{Mixed Domain Periodic Motion (MDPM)} tracker~\cite{islam2017mixed}.

CNN-based diver detection models have recently been investigated for underwater applications as well~\cite{shkurti2017underwater}. Once trained with sufficient data, these models are quite robust to occlusion, noise, and color distortions. Despite the robust performance, the applicability of these models to real-time applications is often limited due to their slow running time. We refer to~\cite{shkurti2017underwater} for a detailed study on the performance and applicability of various deep visual detection models for underwater applications. In this paper, we design a CNN-based model that achieves robust detection performance in addition to ensuring that the real-time operating constraints on board an autonomous underwater robot are met.   

\subsection{Underwater Human-Robot Communication}
Modulating robot motion based on human input in the form of speech, hand gestures, or keyboard interfaces has been explored extensively for terrestrial environments \cite{coronado2017gesture,chen2015hand,wolf2013gesture}. However, most of these human-robot communication modules are not readily applicable in underwater applications due to environmental and operational constraints \cite{dudek2008sensor}. Since visual communication is a feasible and operationally simpler method, a number of visual diver-robot interaction frameworks have been developed in the literature. 

A gesture-based framework for underwater visual servo control was introduced in \cite{dudek2005visually}, where a human operator on the surface was required to interpret the gestures and modulate robot movements. Due to challenging underwater visual conditions \cite{dudek2008sensor} and a lack of robust gesture recognition techniques, fiducial markers were used in lieu of free-form hand gestures as they are efficiently and robustly detectable under noisy conditions. In this regard, the most commonly used fiducial markers have been those with square, black-and-white patterns providing high contrast, such as ARTags~\cite{fiala2005artag} and April-Tags~\cite{olson2011apriltag}, among others. 
Circular markers with similar patterns such as the Photomodeler Coded Targets Module system and Fourier Tags \cite{sattar2007fourier} have also been used in practice.

\begin{figure}[b]
\vspace{-1mm}
    \centering
    \includegraphics[width=0.49\linewidth]{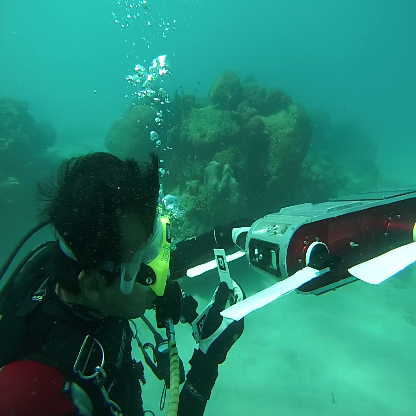} \includegraphics[width=0.49\linewidth]{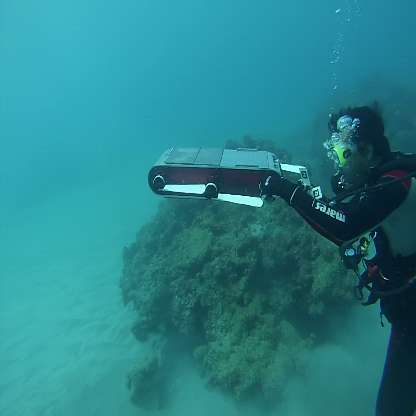}
    \caption{A diver is using ARTags to communicate instructions via the RoboChat language~\cite{dudek2007visual} to an underwater robot during a mission.}
    \label{fig:tag}
    \end{figure}

RoboChat \cite{dudek2007visual} is a visual language proposed for underwater diver-robot communication, for which divers use a set of ARTag markers printed on cards to display predefined sequences of symbolic patterns to the robot (Figure \ref{fig:tag}). These symbol sequences are mapped to commands using a set of grammar rules defined for the language. 
These grammar rules include both terse imperative action commands as well as complex procedural statements. Despite its utility, RoboChat suffers from two critical weaknesses. Firstly, because a separate marker is required for each \emph{token} (\emph{i.e.}, a language component), a large number of marker cards need to be securely carried during the mission, and divers have to search for the cards required to formulate a syntactically correct script; this whole process imposes a rather high cognitive load on the diver. Secondly, the symbol-to-instruction mapping is inherently unintuitive, which makes it inconvenient for rapidly programming a robot. The first limitation is addressed in~\cite{xu2008natural}, in which a set of discrete motions using a pair of fiducial markers is interpreted as a robot command. Different features such as shape, orientation, and size of these gestures are extracted from the observed motion and mapped to the robot instructions. Since more information is embeddable in each trajectory, a large number of instructions can be supported using only two fiducial markers. However, this method introduces additional computational overhead to track the marker motion and needs robust detection of shape, orientation, and size of the motion trajectory. Furthermore, these problems are exacerbated by the fact that both robot and human are suspended in a six-degrees-of-freedom (6DOF) environment. Also, the symbol-to-instruction mapping remains unintuitive.

Since the traditional method for communication between scuba divers is with hand gestures, similarly instructing a robot using hand gestures is more intuitive and flexible than using fiducial markers. There exist a number of hand gesture-based HRI frameworks \cite{coronado2017gesture, chen2015hand,wolf2013gesture} for terrestrial robots. In addition, recent visual hand gesture recognition techniques \cite{molchanov2015hand, neverova2014multi} based on CNNs have been shown to be highly accurate and robust to noise and visual distortions \cite{fabbri2018enhancing}. A number of such visual recognition and tracking techniques have been successfully used for underwater tracking \cite{shkurti2017underwater} and have proven to be more robust than other purely feature-based methods \cite{islam2018person}. However, the feasibility of these models for hand gesture-based human-robot communication has not yet been explored in-depth, which we attempt to do in this paper. In addition, we demonstrate that off-the-shelf deep visual detection models (\emph{e.g.},~\cite{tfzoo}) can be utilized in our framework to ensure robust performance.

\section{Autonomous Diver Following}
In the following sections, we present two methodologies for an underwater robot to visually detect and track a diver. Once the diver is localized in the image space, a visual servoing controller~\cite{espiau1992new} regulates motion commands in six degrees of freedom space in order to follow the diver in a smooth trajectory. We will further discuss the operation of our visual servoing controller in Section~\ref{vizSer}.      

\subsection{Mixed Domain Periodic Motion (MDPM) Tracker}
\begin{figure}[t]
\vspace{1mm}
 \centering
    \includegraphics[width=\linewidth]{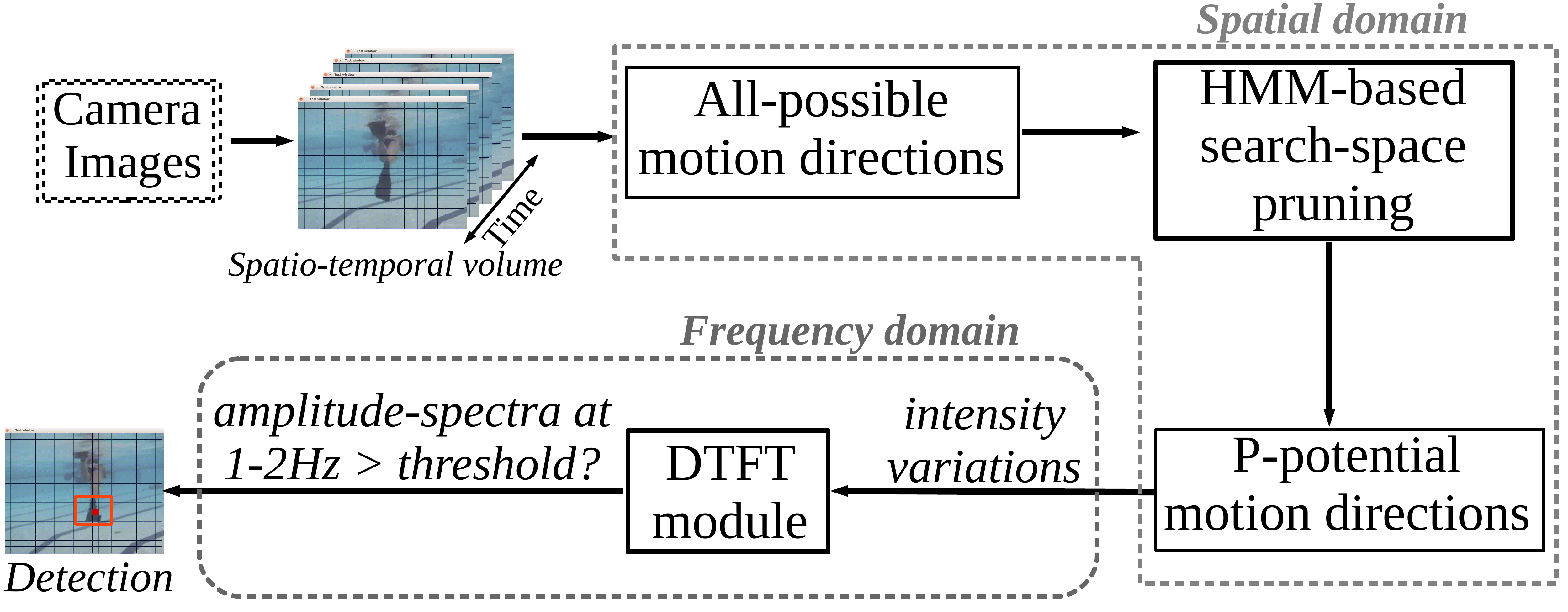}
 \caption{An outline of the MPDM tracker.}
 \vspace{-3mm}
 \label{fig:mdpm}
\end{figure}

MDPM tracker uses both spatial-domain and frequency-domain features to visually track a diver's motion over time. As illustrated in Figure \ref{fig:mdpm}, the overall process can be summarized as follows:   
\begin{itemize}
\item First, the motion direction of a diver is modeled as a sequence of non-overlapping image sub-windows over time, and it is quantified as a vector of intensity values corresponding to those sub-windows. 
\item These captured intensity values (for all possible motion directions) are then exploited by an HMM-based pruning method to discard the motion directions that are unlikely to be the direction where the diver is swimming.
\item Finally, the potentially optimal motion directions are inspected in the frequency-domain. A high amplitude-spectra in the 1-2Hz frequency band is an indicator of a human swimming motion, which is used to locate the diver in the image-space.
\end{itemize}

\vspace{1mm}
\subsubsection{Modeling the Motion Directions of a Diver}
\begin{figure}
\begin{center}
\includegraphics [width=\linewidth]{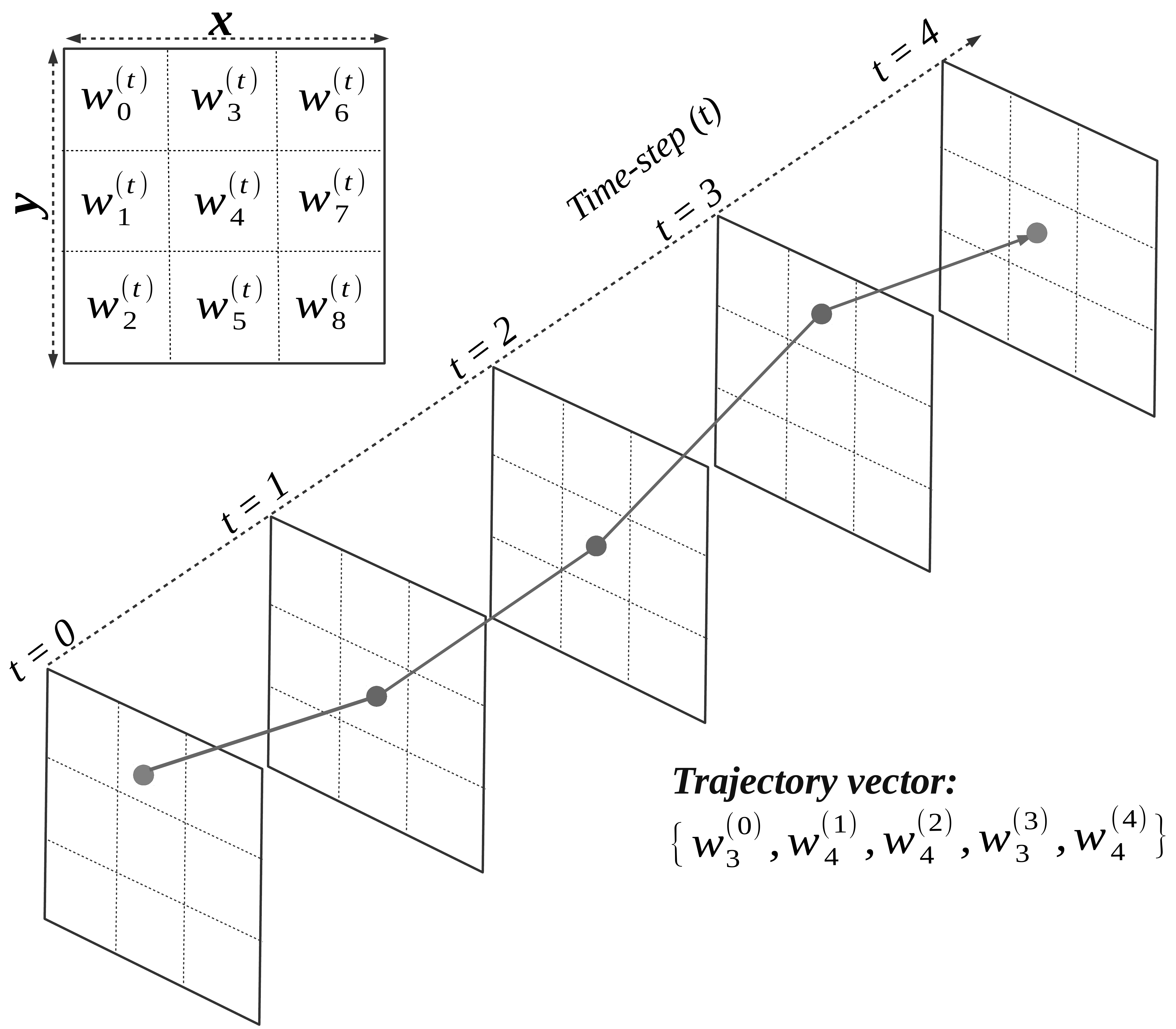}
\caption{A simple scenario with the image-space divided into $M$$=$$9$ windows is shown on the top-left corner. One possible motion direction is shown on the bottom, where the corresponding trajectory vector for $T$$=$$5$ time-steps is $v$ $=$ $\{w^{(0)}_{3}, w^{(1)}_{4}, w^{(2)}_{4}, w^{(3)}_{3}, w^{(4)}_{4}\}$.}
\label{window}
\end{center}
\end{figure}

First, the image-frame at time-step $t$ is divided into a set of $M$ rectangular windows labeled as $w^{(t)}_0, w^{(t)}_1, \dots, w^{(t)}_{M-1}$. Then, the motion directions are quantified as vectors of the form $v$ $=$ $\{ w^{(0)}_{i}, w^{(1)}_{i}, \dots, w^{(t)}_{i}, \dots, w^{(T-1)}_{i}\}$ (Figure \ref{window}). Here, $T$ stands for the \textit{slide-size} and $w^{(t)}_{i}$ denotes one particular window on the $t^{th}$ frame ($i \in [0,M-1]$)  where $t = \{0,1, \dots, (T-1)\}$. We call $v$ the \textit{trajectory vector}.

Now, let $x_v$ denote the \textit{intensity vector}\footnote{We refer to \textit{intensity value} of a window as the Gaussian-filtered average intensity of that window} corresponding to the trajectory vector $v$. We interpret this sequence of $T$ numbers in $x_v$ as values of a discrete aperiodic function defined on $t = 0, 1, \dots, (T-1)$. This interpretation allows us to take the Discrete Time Fourier Transform (DTFT) of $x_v$ and get a $T$-periodic sequence of complex numbers which we denote by $X_{v}$. The values of $X_{v}$ represents the discrete frequency components of $x_v$ in the frequency-domain. The standard equations~\cite{Oppenheim96} that relate the spatial and frequency-domains through a Fourier Transform are as follows:

{\footnotesize
\begin{align}
X_{v} [k] &= \sum_{t=0}^{T-1}{x_v[t] e^{-j 2 \pi t k / N  }} \qquad (k \in [0, N-1])    \\
x_v[t] &= \frac{1}{N} \sum_{k=0}^{N-1}{X_{v} [k] e^{j 2 \pi t k / N  } } 
\label{DFT}
\end{align}
}

As mentioned earlier, we try to capture the periodic motion of the diver in $x_v$ by keeping track of the variations of intensity values along $v$. Then, we take the DTFT of $x_v$ to inspect its amplitude-spectra of the discrete frequency components. The flippers of a human diver typically oscillate at frequencies between $1$ and $2$ Hz \cite{Sattar09RSS}. Hence, our goal is to find the motion direction $v$ for which the corresponding intensity vector $x_v$ produces maximum amplitude-spectra within $1$-$2$Hz in its frequency-domain ($X_{v}$). Therefore, if $\digamma(v)$ is the function that performs DTFT on $x_v$ to generate $X_{v}$ and subsequently finds the amplitude-spectra with high energy responses in the $1$-$2$ Hz range, we can formulate the following optimization problem by predicting the motion direction of a diver as:
\begin{equation}
v^* = \argmax_{v}{\digamma(v)} 
\label{Opti}
\end{equation}

\begin{figure}[h]
\begin{center}
\includegraphics [width=0.8\linewidth]{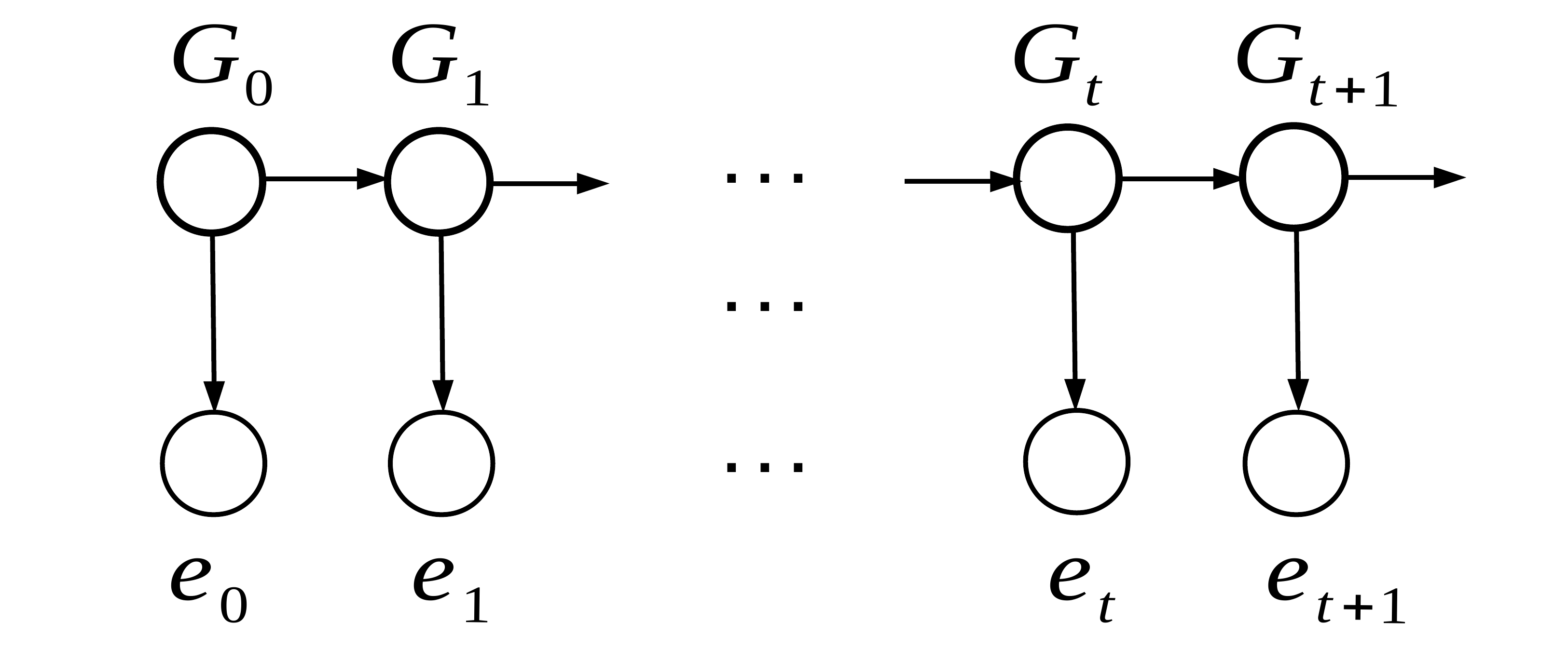}
\caption{An HMM-based representation for the search-space of all possible motion directions. Here, the observed states ($e_t$) represent an \textit{evidence vector} containing intensity values for $w^{(t)}_{i}$ ($i \in [0, M-1]$), whereas the hidden states $G_t$ represent the probabilities that $w^{(t)}_{i}$ contains (a part-of) a diver's flippers.}
\label{HMM}
\end{center}
\vspace{-4mm}
\end{figure} 

The search-space under consideration in optimizing Equation \ref{Opti} is of size $M^T$, as there are $M^T$ different trajectory vectors considering $M$ number of windows and slide-size $T$. Performing $O(M^T)$ computations in a single detection is computationally too expensive for a real-time implementation. Besides, a large portion of all possible motion directions are irrelevant due to the limited body movement capabilities of human divers. Consequently, we adopt a search-space pruning step to eliminate these unfeasible solutions. 

\vspace{1mm}
\subsubsection{HMM-based Search-space Pruning}
We have discussed that the periodic variations of intensity values, being transformed into the frequency-domain, carry information about the swimming direction of the diver. On the other hand, in the spatial-domain, the intensity value (or RGB values) of a particular window suggests whether (a part of) the diver's body or flippers might be present in that window. Therefore, we can assign some degree of confidence \textit{(i.e.}, probability) that the diver is present in a particular window. We do this by first using prior knowledge about the color of the diver's flipper to set an intensity range $R$. We choose $R$ such that the probability of the diver's flipper being present in a window $w^{(t)}_{i}$ at time-step $t$ is given by the following equation: 

\begin{equation}
\footnotesize
P\{ G_{t} = w^{(t)}_{i} | e_t \} \propto \frac{1}{Dist(I(w^{(t)}_{i}), R)}
\end{equation} 

Here, $e_t$ is the \textit{evidence vector} that contains intensity values for window $w^{(t)}_{i}$ ($i \in [0, M-1]$). $Dist(I(w^{(t)}_{i}), R)$ measures the numeric distance between the intensity of window $w^{(t)}_{i}$ and the intensity range $R$. As depicted in Figure \ref{HMM}, we define our HMM structure by considering $G_t$ as a `hidden' state (as we want to predict which window(s) contain(s) the diver's flippers) and $e_t$ as an `observed' state (as we can observe the intensity values of these windows) at time-step $t$. In addition, we consider it unlikely that the diver's flippers will move too far away from a given window in a single time-step. Based on these assumptions, we define the following Markovian transition probabilities:

\begin{equation}
\footnotesize
\begin{gathered}
P\Big\{G_{t+1}=w^{(t+1)}_{i} \Big| G_{0}=w^{(0)}_{i}, G_{1}=w^{(1)}_{i}, \dots, G_{t}=w^{(t)}_{i}  \Big\} \\
\hspace{15mm} = P\Big\{ G_{t+1} = w^{(t+1)}_{i} \Big| G_{t} = w^{(t)}_{i}  \Big\} \\ 
\propto \frac{1}{Dist(w^{(t+1)}_{i}, w^{(t)}_{i} )}
\end{gathered}
\label{Eq:pair}
\end{equation}

\begin{equation}
\footnotesize
P\Big\{e_{t} \Big| G_{t}=w^{(t)}_{i}  \Big\} =
   \begin{cases}
       1-\epsilon &\quad\text{if } I(w^{(t)}_{i}) \in R  \\
        \epsilon &\quad\text{otherwise.} \ 
     \end{cases}
 \label{Eq:evidence}
\end{equation} 

\vspace{1mm}
Here, $Dist(w^{(t+1)}_{i}, w^{(t)}_{i})$ is the Euclidean distance between the centers of window $w^{(t+1)}_{i}$ and $w^{(t)}_{i}$. We take $\epsilon=0.1$ in our implementation. Additionally, as discussed above, we adopted an intensity range $R$ to define $P\{ G_{t} = w^{(t)}_{i} | e_t \}$; color-based ranges (in RGB-space or HSV-space) can also be adopted for this purpose. One advantage of using intensity range is that the intensity values of each window are already available in the trajectory vector and therefore no additional computation is required.

We use this HMM-based setup to predict \textit{the most likely sequence of states} ($G_{0}, \dots G_{T-1}$) that leads to a given state $G_{T}=w^{(T)}_{i}$ at time-step $t$. In terms of the parameters and notations mentioned above, this is defined as follows:

\begin{equation}
\footnotesize
\begin{aligned}
\mu ^*(T) &= \argmax_{w^{(0)}_{i}, \dots, w^{(T-1)}_{i}}{ P \Big\{ G_{0}=w^{(0)}_{i}, \dots, G_{T}=w^{(T)}_{i} \Big| e_{0}, \dots, e_{T}  \Big\}} \\
   &= \argmax_{ w^{(0:T-1)}_{i} }{ P \Big\{ G_{0:T}=w^{(0:T)}_{i} \Big| e_{0:T}  \Big\}}
\end{aligned}
\end{equation}

Here, we adopted the short-form notations in the second line for convenience. Now, using the properties of the Bayesian chain rule and Markovian transition~\cite{rabiner1989tutorial}, a recursive definition of $\mu ^*(T)$ can be obtained as follows (see Appendix I for the derivation):

\begin{equation}
\footnotesize
\begin{aligned}
\mu ^*(T) &= P \Big\{ e_T \Big| G_{T}=w^{(T)}_{i} \Big\} \\ & \times { \argmax_{w^{(T-1)}_{i}}{ \Big( P\Big\{ G_{T}=w^{(T)}_{i} \Big| G_{T-1}=w^{(T-1)}_{i}  \Big\} } } \times { \mu ^*(T-1)  \Big)  }
\end{aligned}
\end{equation}
\label{Recur}

Using this recursive definition of $\mu ^*(T)$, we can efficiently keep track of the most likely sequence of states over $T$ time-steps. This sequence of states corresponds to a sequence of windows, which is effectively the desired trajectory vector. However, a pool of such trajectory vectors is needed so that we can inspect the frequency responses to choose the one having the strongest response. Therefore, we choose the $p$ most likely sequences of states, which we define as $\mu^{*}(T,p)$. Here, $p$ is the \textit{pool-size}. Finally, we rewrite the problem definition in Equation \ref{Opti} as follows:

\begin{equation}
v^* = \argmax_{\mu \in \mu^{*}(T,p)}{\digamma(\mu)} 
\label{FinalOpti}
\vspace{2mm}
\end{equation}

The procedure for finding $v^*$ is outlined in Appendix II. Here, at each detection cycle, we first find the $p$ most potential motion directions (\textit{i.e.}, trajectory vectors) through the HMM-based pruning mechanism. We do this efficiently using the notion of \textit{dynamic programming}. As evident from the algorithm, it requires $\mathcal{O}(M^2)$ operations to update the dynamic table of probabilities at each detection cycle. 

\vspace{2mm}
\subsubsection{Frequency-Domain Detection}
Once the potential trajectory vectors are found, we perform DTFT to observe their frequency-domain responses. The trajectory vector producing the highest amplitude-spectra at $1$-$2$Hz frequencies is selected as the optimal solution. DTFT can be performed very efficiently; for instance, the running-time of a Fast Fourier Transform algorithm is $\mathcal{O}(T\times logT)$, where $T$ is the size of the input vectors. Therefore, we need only $\mathcal{O}(p \times T\times logT)$ operations for inspecting all potential trajectory vectors. Additionally, the approximated location of the diver is readily available in the solution; therefore, no additional computation is required for tracking.

\subsection{A CNN-based Model for Diver Detection}\label{sec:cnnDF}
\begin{figure}[h]
\begin{center}
\includegraphics [width=\linewidth]{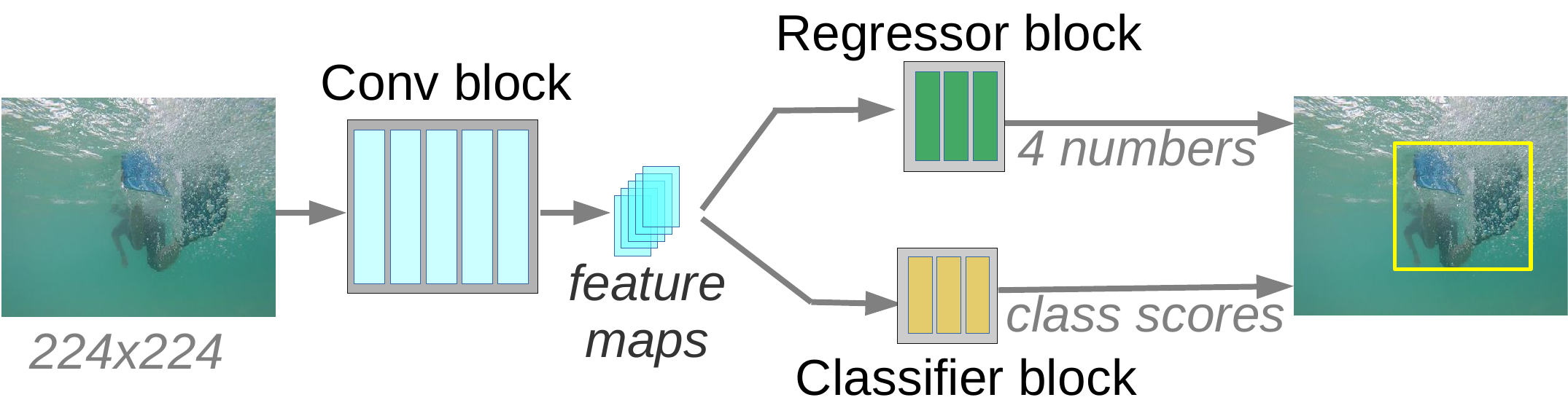}
\caption{A schematic diagram of our CNN-based model for detecting a single diver in RGB image-space.}
\label{dr_detect}
\end{center}
\end{figure}

A major limitation of MDPM tracker is that it does not model the appearance of a diver, it only detects the periodic signals pertaining to a diver's flipping motion. In addition, its performance is affected by the diver's swimming trajectory (straight-on, sideways, etc.), the color of wearables, etc. We try to address these issues and ensure robust detection performance by using a CNN-based model for diver detection. Figure \ref{dr_detect} illustrates a schematic diagram of the model. It consists of three major parts: a convolutional block, a regressor block, and a classifier block.  

The convolutional block consists of five layers, which extracts the spatial features in the RGB image-space by learning a set of convolutional kernels. The extracted features are then fed to classifier and regressor blocks for detecting a diver and localizing the corresponding bounding box, respectively. Both the classifier and regressor blocks consist of three fully connected layers. In our implementation, we have three object categories: diver, robot, and background. Therefore, the regressor block learns to detect a diver or a robot in an image by extracting the background, whereas the classifier block learns the objectness scores (\textit{i.e.}, confidence) associated with those detections. 

\begin{table}[h]
\centering
\caption{Parameters and dimensions of the CNN model outlined in Figure \ref{dr_detect}. (convolutional block: conv1-conv5, classifier block: fc1-fc3, regression block: rc1-rc3; n: the number of object categories; *an additional pooling layer was used before passing the conv5 features-maps to fc1)}
\footnotesize
\begin{tabular}{|m{1cm} m{1.55cm} m{1.55cm} cm{1.5cm}|}
\hline
Layer & Input feature-map & Kernel size & Strides  & Output feature-map \\ \hline \hline
conv1 & 224x224x3 & 11x11x3x64 & [1,4,4,1]   & 56x56x64 \\
pool1 & 56x56x64 & 1x3x3x1 & [1,2,2,1]  & 27x27x64  \\  \hline
conv2 & 27x27x64 & 5x5x64x192 & [1,1,1,1]   & 27x27x192 \\
pool2 & 27x27x192 & 1x3x3x1 & [1,2,2,1]  &  13x13x192  \\ \hline 
conv3 & 13x13x192 & 3x3x192x192 & [1,1,1,1]   & 13x13x192  \\  
conv4 & 13x13x192 & 3x3x192x192 & [1,1,1,1]   & 13x13x192  \\  
conv5 & 13x13x192 & 3x3x192x128 & [1,1,1,1]   & 13x13x128  \\ \hline \hline 
fc1 & 4608x1$^*$ & $-$ & $-$   & 1024x1  \\
fc2 & 1024x1 & $-$ & $-$ & 128x1  \\
fc3 & 128x1 & $-$ & $-$   & n  \\ \hline \hline
rc1 & 21632x1 & $-$ & $-$ & 4096x1  \\
rc2 & 4096x1 & $-$ & $-$ & 192x1  \\
rc3 & 192x1 & $-$ & $-$ & 4n  \\ \hline 
\end{tabular}
\label{tab:conv}
\end{table}

The main reason for designing such a simple model is the computational overhead. Our objective is to design a robust detector that also ensures real-time performance in an embedded platform. The state-of-the-art deep visual detectors often use region proposal networks and dense models that are computationally demanding \cite{tfzoo}. Therefore, we do not use off-the-shelf models and choose to design this simpler model for diver detection. We will present the training process and other operational details in Section \ref{sec:Perf}.

\begin{figure*}[t]
\vspace{1mm}
\centering
\includegraphics [width=\linewidth]{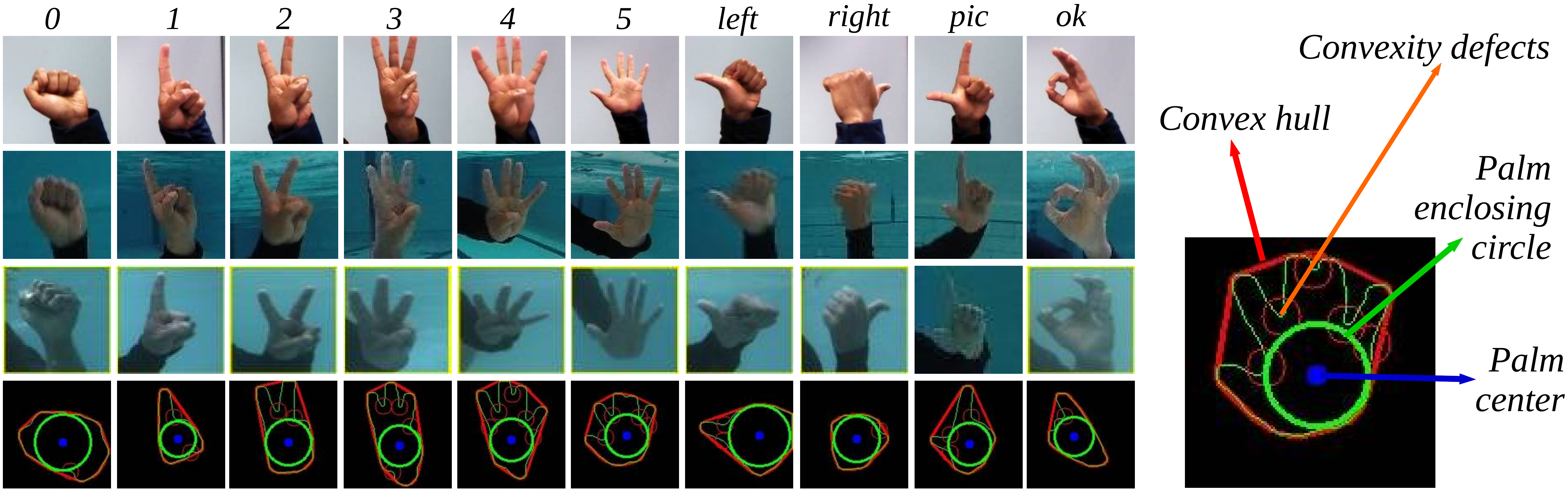}
\vspace{-4mm}
\caption{The first three rows on the left show a few sample training images for the ten classes of hand gestures used in our framework; the bottom row shows the expected hand-contours with different curvature markers for each class of gestures. The annotated curvature markers for a particular example are shown on the right.}
\label{data}
\end{figure*}

\begin{figure}
\centering
\includegraphics [width=\linewidth]{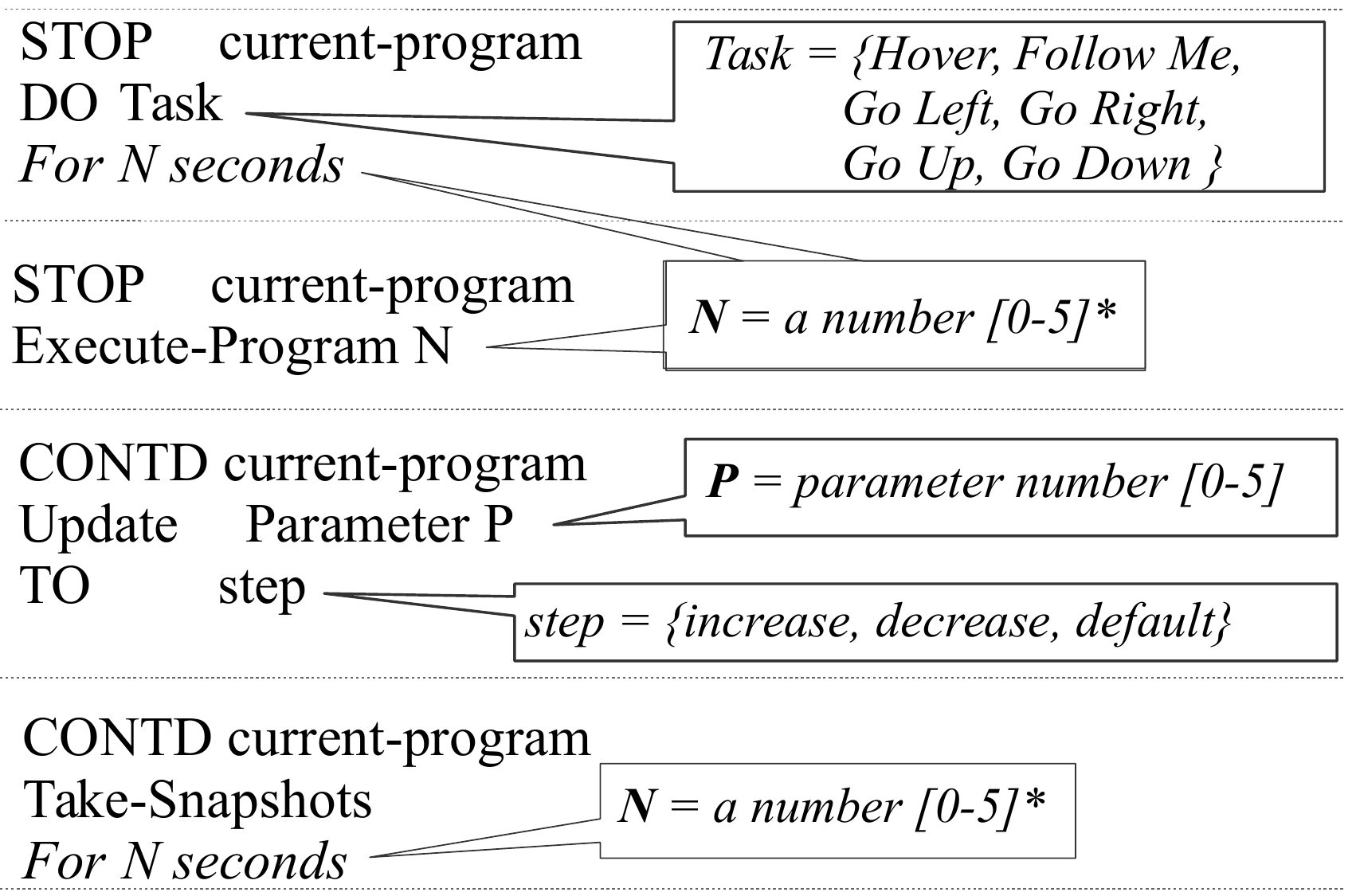}
\vspace{-4mm}
\caption{A set of \textit{task switching} and \textit{parameter reconfiguration} instructions that are currently supported by our framework.}
\label{ins}
\end{figure}

\vspace{2mm}
\section{Human-Robot Communication}\label{sec:RoboChatgest}
Our proposed framework is built on a number of components:  the choice of hand gestures to map to instruction tokens, the robust recognition of hand gestures, and the use of a finite-state machine to enforce the instruction structure and ignore erroneous detections or malformed instructions. Each of these components is described in detail in the following sections.

\subsection{Mapping Hand Gestures to Instruction Tokens}
\label{sec:gesture-to-token}
Our objective is to design a simple yet expressive framework that can be easily interpreted and adopted by divers for communicating with the robot without memorizing complex language rules. Therefore, we choose a small collection of visually distinctive and intuitive gestures, which would improve the likelihood of robust recognition in degraded visual conditions. Specifically, we use only the ten gestures shown in Figure~\ref{data}; as seen in this figure, each gesture is intuitively associated with the command it delivers. Sequences of different combinations of these gestures formed with both hands are mapped to specific instructions. As illustrated in Figure~\ref{ins}, we concentrate on the following two sets of instructions in our framework: 

\begin{itemize}
\item \textbf{Task switching}: This is to instruct the robot to stop the execution of the current program and start a new task specified by the diver, such as hovering, following, or moving left/right/up/down, etc. These commands are atomic behaviors that the robot is capable of executing. An optional argument can be provided to specify the duration of the new task (in seconds). An operational requirement is that the desired programs need to be numbered and known to the robot beforehand.

\item \textbf{Parameter reconfiguration}: This is to instruct the robot to continue the current program with updated parameter values. This enables underwater missions to continue unimpeded (as discussed in Section~\ref{sec:intro}), without interrupting the current task or requiring the robot to be brought to the surface. Here, the requirement is that the tunable parameters need to be numbered and their choice of values need to be specified beforehand. The robot can also be instructed to take pictures (for some time) while executing the current program. 
\end{itemize}

\begin{figure}
\vspace{1mm}
\centering
\includegraphics [width=\linewidth]{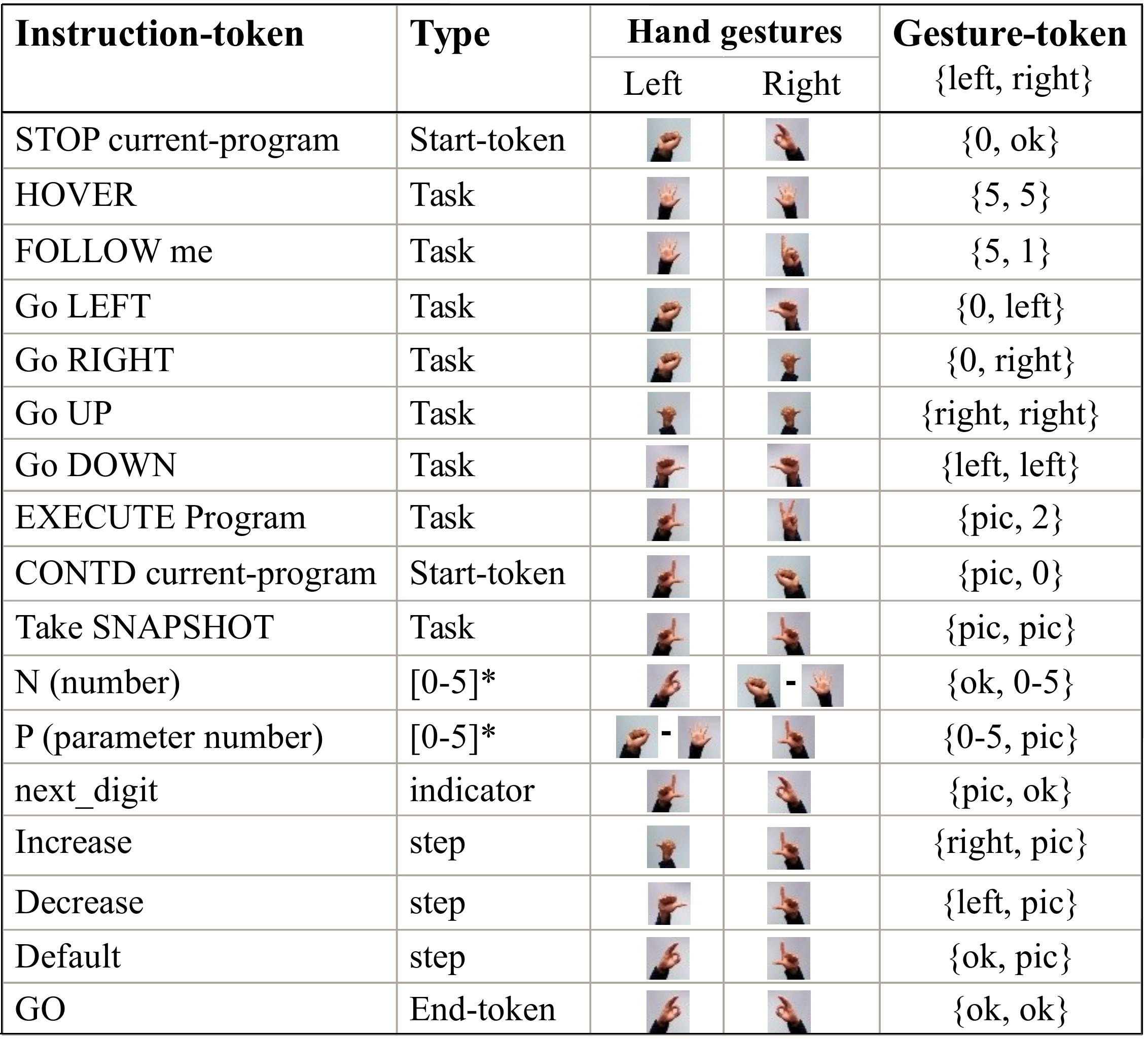}
\vspace{-5mm}
\caption{The mapping of gesture-tokens to instruction-tokens used in our framework.}
\label{ins_map}
\vspace{-3mm}
\end{figure}

The proposed framework supports a number of task switching and parameter reconfiguration instructions, which can be extended to accommodate more instructions by simply changing or appending a user-editable configuration file. The hand gesture-to-token mapping is carefully designed so that the robot formulates executable instructions only when intended by the diver. This is done by attributing specific hand gestures as \emph{sentinels} (\emph{i.e.}, \textit{start-} or \textit{end-tokens}). Figure~\ref{ins_map} illustrates the gesture to atomic-instruction mapping used in our framework. Additional examples are shown in Appendix III, where a series of $(start\_token, instruction, end\_token)$ tuples are mapped to their corresponding sequences of $gesture\_tokens$.

\subsection{Hand Gesture Recognition and Instruction Generation}\label{HandGest}
\begin{figure}[h]
\centering
\includegraphics [width=\linewidth]{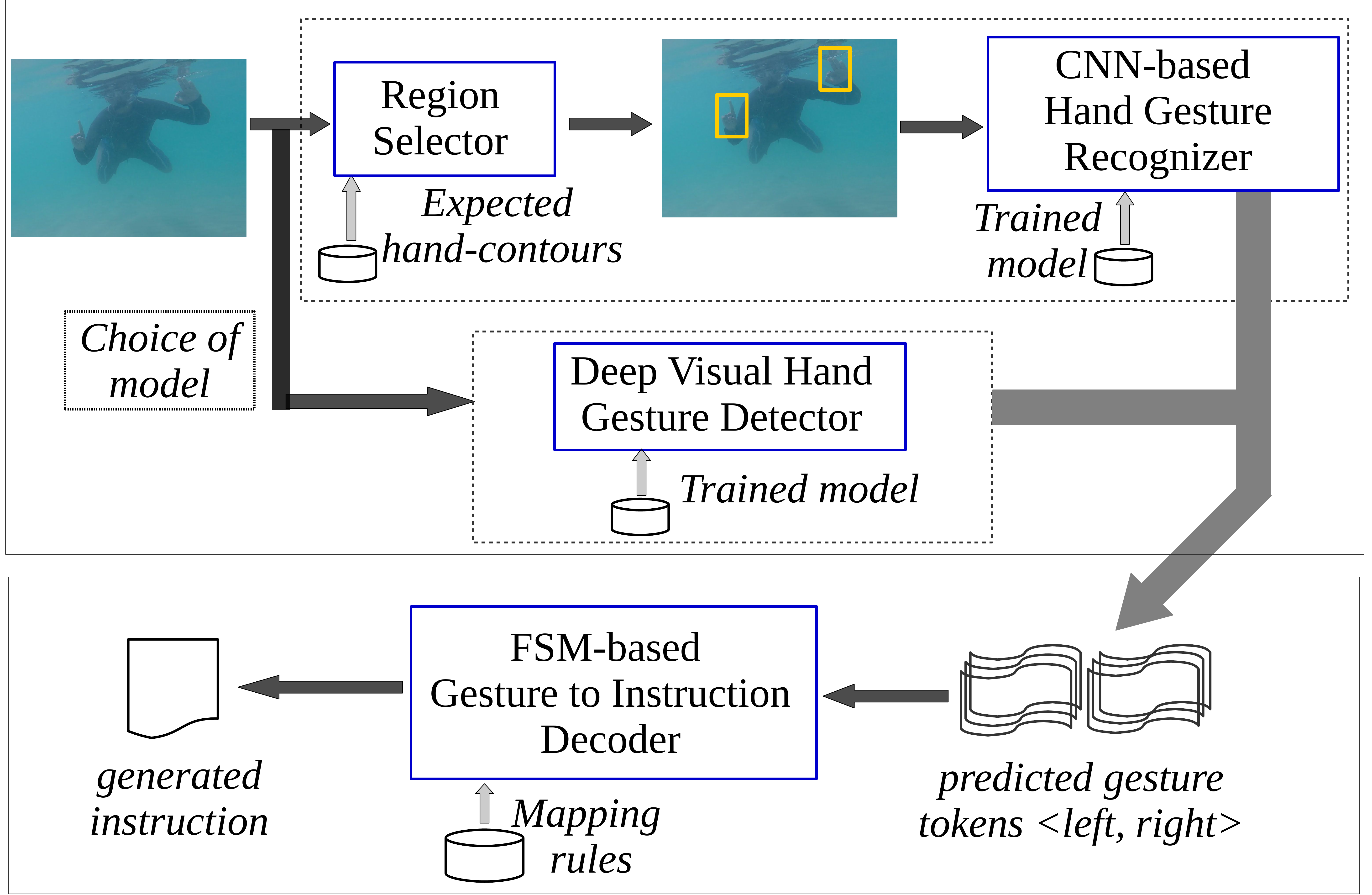}
\vspace{-4mm}
\caption{An overview of the process of mapping hand gestures to instructions in our framework. The top block demonstrates two (choices of) hand gesture recognition systems, and the bottom block depicts a finite-state machine for hand gesture-to-instruction mapping. 
}
\vspace{-2mm}
\label{Proc}
\end{figure}

Robust mapping of gesture-tokens to instruction-tokens is essential for a human-robot communication system in general. As illustrated in Figure ~\ref{Proc}, in our proposed framework, the challenges lie in localizing the hand gestures in the image-space, accurately recognizing those hand gestures, and then mapping them to the correct instruction-tokens. We now provide the implementation details of these components in the following sections.  

\vspace{1mm}
\subsubsection{Region Selection} \label{sec:reg}
To detect gestures, the hand regions need to be cleanly extracted from the image. The CNN-based region proposal networks~\cite{renNIPS15fasterrcnn} or classical methods such as Edge-box~\cite{zitnick2014edge} are known to be robust and highly accurate in segmenting prospective regions for object detection. However, due to their slow running time in embedded platforms, we adopt the classical image processing techniques to select prospective hand regions in the image-space. As illustrated in Figure \ref{reg}, the overall region selection process can be summarized as follows:

\begin{figure}[h]
\centering
\includegraphics [width=\linewidth]{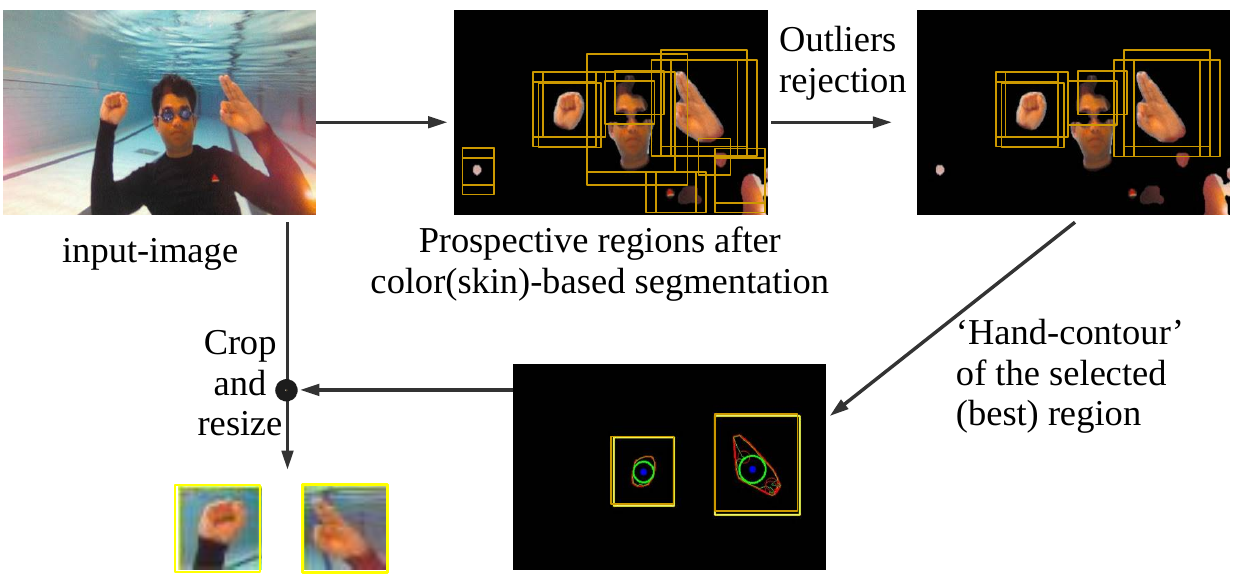}
\vspace{1mm}
\caption{Outline of the region selection mechanism of our framework: first, the (skin) color-based segmentation is performed to get potential regions for hand gestures; then, the outlier regions are discarded based on cached information about the previous locations of the hands.}
\label{reg}
\vspace{-2mm}
\end{figure} 

\begin{enumerate}[i.]
\item First, the camera image (RGB) is blurred using Gaussian smoothing and then thresholded in the HSV space for skin-color segmentation \cite{oliveira2009skin}. We assume that the diver performs gestures with bare hands; if the diver is to wear gloves, the color thresholding range in the HSV space needs to be adjusted accordingly.
\vspace{1mm}
\item Contours of the different segmented regions in the filtered image space are then extracted (see Figure~\ref{data}). Subsequently, different contour properties such as convex hull boundary and center, convexity defects, and important curvature points are extracted. We refer readers to \cite{yeo2015hand} for details about the properties and significance of these contour properties.

\item Next, the outlier regions are rejected using cached information about the scale and location of hand gestures detected in the previous frame. This step is, of course, subject to the availability of the cached information.
\vspace{1mm}
\item Finally, the hand contours of potential regions are matched with a bank of hand contours that are extracted from training data (one for each class of hand gestures as shown in the bottom row of Figure \ref{data}). The final regions for left- and right-hand gestures are selected using the proximity values of the closest contour match \cite{yeo2015hand} (\textit{i.e.}, the region that is most likely to contain a hand gesture is selected).     
\end{enumerate}

\begin{figure*}[t]
\vspace{1mm}
\centering
\includegraphics [width=\linewidth]{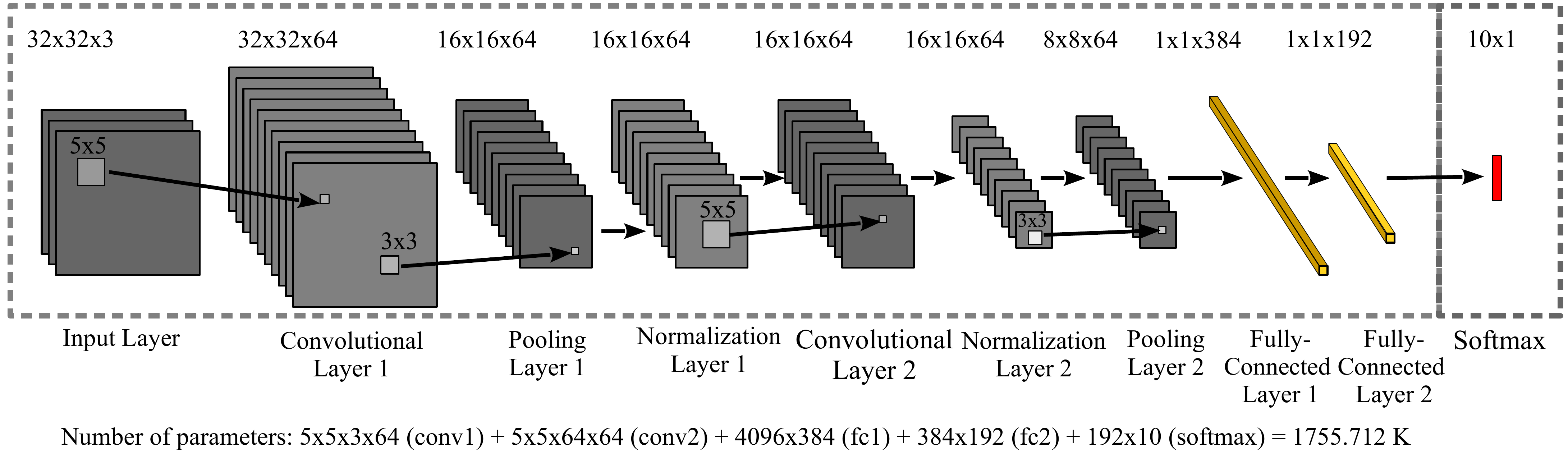}
\vspace{-5mm}
\caption{Architecture of the CNN model used in our framework for hand gesture recognition.}
\label{Arch}
\end{figure*} 

\vspace{1mm}
\subsubsection{CNN Model for Gesture Recognition} 
Following region selection, the cropped and resized $32\times 32$ image-patches are fed to a CNN-based model for hand gesture recognition. The architecture of the model is illustrated in Figure~\ref{Arch}. Two convolutional layers are used for extracting and learning the spatial information within the images. Spatial down-sampling is done by max-pooling, while the normalization layer is used for scaling and re-centering the data before feeding it to the next layer. The extracted feature-maps are then fed to the fully connected layers to learn decision hyperplanes within the distribution of training data. Finally, a soft-max layer provides the output probabilities for each class, given the input data. Note that similar CNN models are known to perform well for small-scale (\textit{i.e.}, $10$-class classification) problems which are similar to ours. The dimensions of each layer and associated hyper-parameters are specified in Figure \ref{Arch}; details about the training process will be provided in Section \ref{sec:Perf}. 

\begin{figure*}[h]
\vspace{1mm}
\centering
\includegraphics [width=\linewidth]{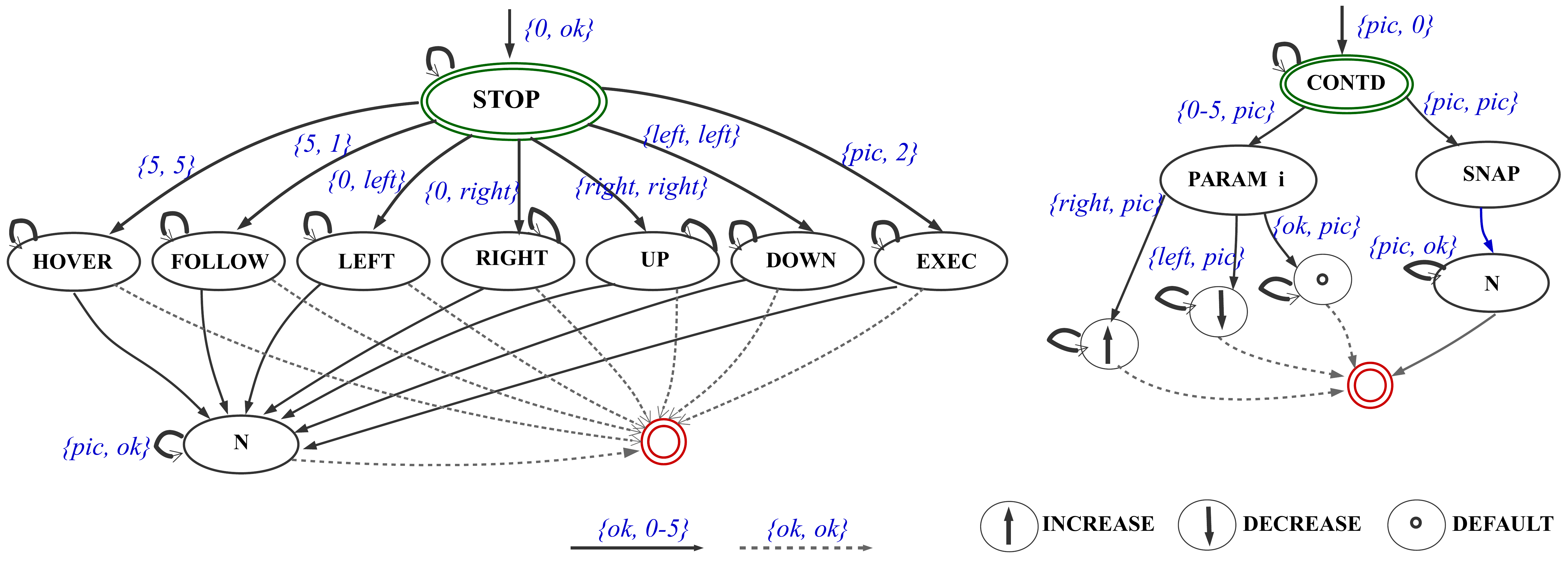}
\vspace{-4mm}
\caption{An FSM-based deterministic mapping of hand gestures to instructions (based on the rules defined in Figure \ref{ins_map}).}
\label{Fsm}
\end{figure*} 

\begin{figure*}[h]
\vspace{1mm}
    \centering
    \begin{subfigure}[t]{\textwidth}
        \includegraphics[width=\linewidth]{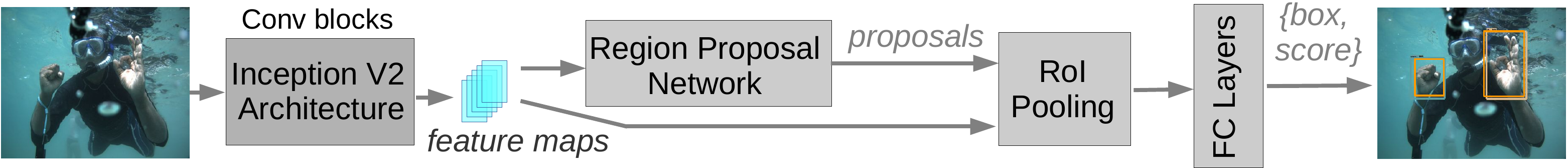}
        \caption{Faster RCNN~\cite{renNIPS15fasterrcnn} with inception V2~\cite{szegedy2016rethinking}}
    \end{subfigure}
    
    \vspace{3mm}
    \begin{subfigure}[t]{\textwidth}
        \includegraphics[width=\linewidth]{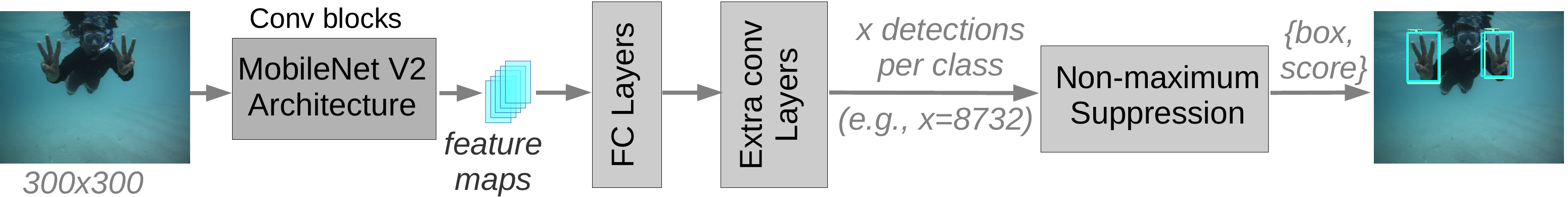}
        \caption{SSD~\cite{liu2016ssd} with MobileNet V2~\cite{sandler2018inverted}}
    \end{subfigure} 
   
 \caption{A schematic diagram of the two deep visual detectors used in our framework for hand gesture recognition.}
 \label{fig:deepGest}
\end{figure*}

\subsubsection{Deep Visual Detectors for Hand Gesture Recognition}
One operational convenience of hand gesture-based programming is that the robot stays in `hover' mode during the process, and the overall operation is not as time-critical as in the diver following scenario. Therefore, we investigate if we could use deeper and denser models to improve the robustness and accuracy of hand gesture recognition by sacrificing its running time. Specifically, we explore the applicabilities of the state-of-the-art deep visual models for hand gesture recognition and try to balance the trade-offs between accuracy and running time. 

We use two fast object detectors \cite{tfzoo}: Faster RCNN with Inception v2 and Single Shot MultiBox Detector (SSD) with MobileNet v2. As illustrated in Figure \ref{fig:deepGest}, they are end-to-end models, \textit{i.e.}, they perform region selection and hand gesture classification in a single pass. Additionally, they are known to provide highly accurate and robust performances in noisy visual conditions. 

\begin{enumerate}[i.]
\item Faster RCNN with Inception v2: Faster RCNN~\cite{renNIPS15fasterrcnn} is an improvement of R-CNN~\cite{Girshick2014RCNN_CVPR} that introduces a Region Proposal Network (RPN) to make the whole object detection network end-to-end trainable. The RPN uses the last convolutional feature-maps to produce region proposals, which is then fed to the fully connected layers for the final detection. The original implementation uses VGG-16 \cite{simonyan2014very} for feature extraction; we use the Inception v2 \cite{szegedy2016rethinking} as the feature extractor instead because it is known to produce better object detection performance in standard datasets \cite{tfzoo}. 
\vspace{1mm}
\item SSD with MobileNet v2: 
SSD~\cite{liu2016ssd} is another object detection model that performs object localization and classification in a single pass of the network. However, it does not use an RPN; it uses the regression trick introduced in the You Only Look Once (YOLO) \cite{redmon2016yolo9000} model. The architectural difference of SSD with YOLO is that it introduces additional convolutional layers to the end of a base network, which results in an improved performance. In our implementation, we use MobileNet v2 \cite{sandler2018inverted} as the base network. 
\end{enumerate}

\subsection{FSM-based Gesture to Instruction Decoder}
An FSM-based deterministic model is used in our model for efficient gesture-to-instruction mapping. As illustrated in Figure~\ref{Fsm}, the transitions between the instruction-tokens are defined as functions of gesture-tokens based on the rules defined in Figure \ref{ins_map}. Here, we impose an additional constraint that each gesture-token has to be detected for $10$ consecutive frames for the transition to be activated. This constraint adds robustness to prevent missed or wrong classification for a particular gesture-token. Additionally, it helps to discard noisy tokens which may be detected when the diver changes from one hand gesture to the next. Furthermore, since the mapping is one-to-one, it is highly unlikely that a wrong instruction will be generated even if the diver mistakenly performs some inaccurate gestures because there are no transition rules other than the correct ones at each state.
\begin{figure*}
\vspace{1mm}
    \centering
    \begin{subfigure}[t]{0.51\textwidth}
        \includegraphics[width=\linewidth]{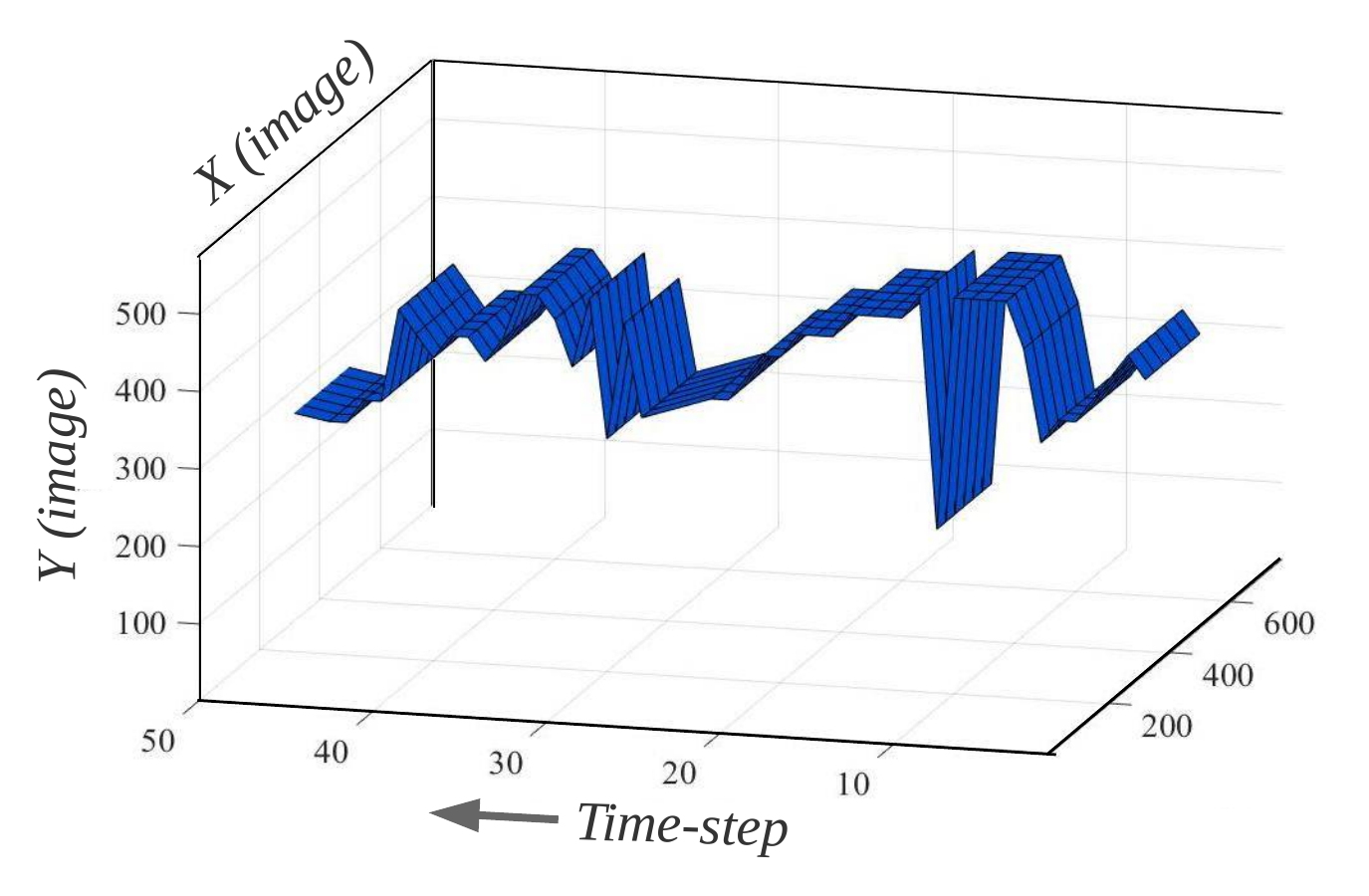}
        \caption{The swimming trajectory of a diver is visualized using a surface-plot; it is prepared off-line by projecting the detected trajectory vectors to the spatio-temporal volume (for a closed-water experiment with $50$ seconds of swimming). }
    \end{subfigure}
    ~
    \begin{subfigure}[t]{0.46\textwidth}
        \includegraphics[width=\linewidth]{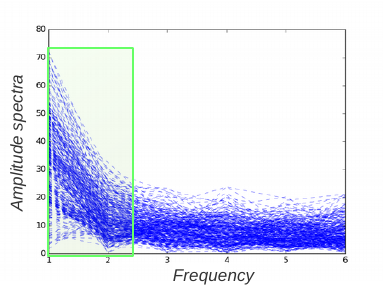}
        \caption{Corresponding frequency-domain signatures are shown; each dotted line represents the amplitude spectra for a single detection in the low-frequency bands (with a sliding window size of $15$).}
    \end{subfigure} 
    \vspace{3mm}
    
    \begin{subfigure}[t]{\textwidth}
        \includegraphics[width=0.99\linewidth]{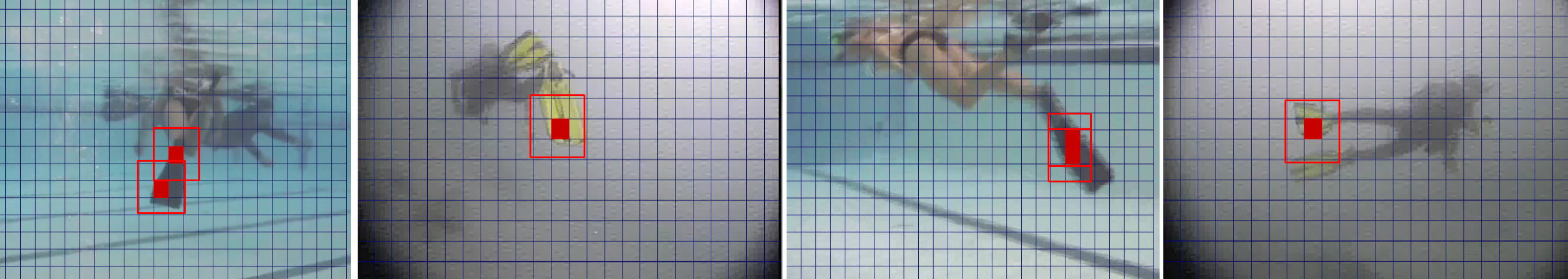}
        \caption{A few snapshots showing the detection of a diver's flipping motion in different scenarios: swimming straight-on away from the robot and swimming sideways (both in closed-water and open-water conditions).}
    \end{subfigure}
    \vspace{1mm}
 \caption{Experimental results for autonomous diver following using the MDPM tracker.}
 \label{fig:mdpm_res}
\end{figure*}

\vspace{3mm}
\section{EXPERIMENTS}\label{sec:Perf}
We now discuss the implementation details of the proposed methodologies and present the experimental results.

\subsection{Experimental Setup}\label{sec:setup}
We have performed several real-world experiments both in closed-water and in open-water conditions (\textit{i.e.}, in pools and in oceans). 
Two underwater robots are used in our experiments and for data collection: an autonomous robot of the Aqua~\cite{dudek2007aqua} family, and an OpenROV~\cite{OpenROV} underwater drone. Both the robots are used for data collection; however, only the Aqua is used for actual experiments since OpenROV does not have an interface for autonomous on-board computation.  

In the diver following experiments, a diver swims in front of the robot in arbitrary directions. The task of the robot is to visually detect the diver using its camera feed and follow behind him/her with a smooth motion. In the hand gesture recognition experiments, a diver faces the robot's camera and performs hand gestures to communicate various instructions to the robot. In this case, the robot's task is to successfully detect and execute the specified instructions. 

\vspace{1mm}
\subsubsection{Training Process for the Deep Models}
We train our supervised deep models on several datasets of hand-annotated images. These images are collected during our field trials and other underwater experiments. We use a Linux machine with four GPU cards (NVIDIA GTX 1080) for the training purposes. Once the training is done, the trained inference model is saved and transferred to the robot CPU for validation and real-time experiments.

\vspace{1mm}
\subsubsection{Visual Servoing Controller}\label{vizSer}
The Aqua robots have a five degree-of-freedom control, \textit{i.e.}, three angular (yaw, pitch, and roll), and two linear (forward and vertical speed) controls. In our experiments for autonomous diver following, we adopt a tracking-by-detection method where the visual servoing~\cite{espiau1992new} controller uses the uncalibrated camera feeds for navigation. The controller regulates the motion of the robot in order to bring the observed bounding box of the target diver to the center of the camera image. The distance of the diver is approximated by the size of the bounding box and forward velocity rates are generated accordingly. Additionally, the yaw and pitch commands are normalized based on the horizontal and vertical displacements of the observed bounding box-center from the image-center; these navigation commands are then regulated by four separate PID controllers. On the other hand, the roll stabilization and hovering are handled by the robot's autopilot module~\cite{meger20143d}. 

\begin{table}
\footnotesize
\caption{Detection performances of MDPM tracker in different swimming conditions.}
  \begin{tabular}{|p{1.1cm}|p{1.35cm}|p{1.35cm}|p{1.35cm}|p{1.35cm}|}
  \hline
    \multirow{2}{*}{Cases} &
      \multicolumn{2}{c|}{Closed Water} &
      \multicolumn{2}{c|}{Open Water} \\
      \cline{2-5}
      & Straight-on & Sideways & Straight-on & Sideways \\  
       \hline \hline
    Positive detection  & \textbf{$647$  ($91.7\%$)} & \textbf{$463$ ($87.3\%$)} & \textbf{$294$ ($85.2\%$)} & \textbf{$240$ ($84.2\%$) } \\
    \hline
    Missed detection  & $46$ ($6.5\%$) & $57$ ($10.8\%$) & $38$ ($11\%$) & $43$ ($15\%$)  \\
    \hline
    Wrong detection & $12$ ($1.8\%$) & $10$ ($1.9\%$) & $13$ ($3.8\%$) & $2$ ($0.8\%$)  \\ \hline
  \end{tabular}
\label{MDPM_com} 
\vspace{-1mm}
\end{table}

\subsection{Results for Autonomous Diver Following}\label{sec:DF}
In our implementation, a monocular camera feed is used by the diver-following algorithms to visually detect a diver in the image-space and generate a bounding box. The visual servoing controller uses this bounding box to regulate robot motion commands in order to follow the diver. Therefore, correct detection of the diver is essential for overall success of the operation. In the following sections, we discuss the detection performances of the two proposed diver-following methodologies.  

\vspace{1mm}
\subsubsection{Implementation and Performance Evaluation of the MDPM Tracker}\label{sec:mdpmDF}
The MDPM tracker has three hyper-parameters: the slide-size ($T$), the size of the sub-windows, and the amplitude threshold ($\delta$) in the frequency-domain. We empirically determine their values through extensive simulations on video footages of diver-following. We found that $T$$=$$15$ and a sub-window size of $30$$\times$$30$ work well in practice; also, we set the frequency threshold $\delta$$=$$75$. We refer to \cite{islam2017mixed} for the experimental details on how these hyper-parameters are chosen.

Once the bootstrapping is done (with the first $T$ frames), mixed-domain detection is performed at every $T$ frames onward in a sliding-window fashion. At each detection, the tracker estimates the potential trajectory vectors that represent a set of motion directions in spatio-temporal volume. If a potential motion direction produces amplitude-spectra more than $\delta$, it is reported as a positive detection at that time-step. Subsequently, the diver's flippers are located in the image-space, and a bounding box is generated. 

Figure \ref{fig:mdpm_res} demonstrates how MDPM tracker detects a diver using spatial- and frequency-domain cues. It keeps track of the diver's motion direction through a sequence of $30$$\times$$30$$\times$$15$ sub-windows in the spatio-temporal volume. The corresponding surface through the image-space over time mimics the actual motion direction of the diver, which indicates the effectiveness of the algorithm.

Table \ref{MDPM_com} provides the performance of MDPM tracker in terms of positive detections, missed detections, and wrong detections for different experimental cases. It achieves a positive detection accuracy of $84.2$-$91.7\%$, which suggests that it provides $8$-$9$ positive detections of a diver per second (considering a frame-rate of $10$ fps). We have found this detection rate quite sufficient for successfully following a diver in practice. 

\begin{figure*}[t]
\vspace{1mm}
    \centering
\includegraphics[width=\linewidth]{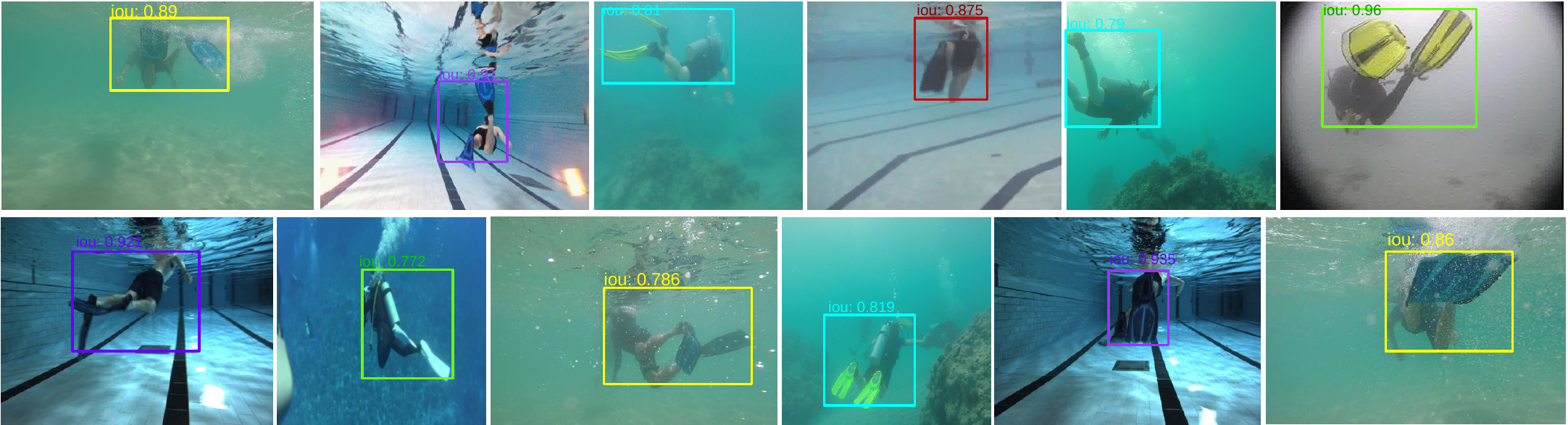}
\vspace{1mm}
 \caption{Detection performance of our CNN-based model on a few images in the validation set; the average accuracy and average IOU is observed to be $0.972$ and $0.674$, respectively.}
 \label{fig:diverDeep}
\end{figure*}

\vspace{1mm}
\subsubsection{Implementation and Performance Evaluation of the CNN-based Model}\label{sec:deppDF}
We trained our CNN-based model (Section \ref{sec:cnnDF}) using a dataset of underwater images that are collected during diver-following experiments in several field trials (in pools and in oceans). In our implementation, we consider three classes: divers, robots, and the background. The dataset has over $10$K images per class, and images are annotated to have class-labels and bounding boxes. 

We presented the model architecture in Table \ref{tab:conv}. Several important parameter choices (such as the kernel sizes in different layers) are standard for feature extraction and widely used in the literature, while other hyper-parameters are chosen empirically. Non-supervised pre-training and drop-outs are not used while training. RMSProp \cite{tieleman2012lecture} is used as the optimization function with an initial learning rate of $0.001$. In addition, standard cross-entropy and $L_2$ loss functions are used by the classifier and regressor, respectively. The model is implemented using TensorFlow \cite{abadi2016tensorflow} libraries in Python.

Figure \ref{fig:diverDeep} shows detection performances of the trained model on few images in the validation set. The model is trained for $300$ epochs with a batch-size of $16$; we refer to Appendix IV for visualizing its convergence behavior. The results in Figure \ref{fig:diverDeep} suggest that the detected bounding boxes are mostly accurate. In addition, its performance does not depend on diver's style of swimming, color of flippers or the flipping motion. Therefore, this model is generally applicable in practical scenarios. 

\begin{table}[h]
\footnotesize
\caption{Performances of the trained CNN model for diver detection.}
  \begin{tabular}{|m{1.3cm}|m{1.25cm}|m{1.25cm}|m{1.1cm}|m{1.5cm}|}
  \hline
  Positive Detections & Missed Detections & Wrong Detections & Avg. IOU & FPS (robot CPU) \\ \hline \hline
  $97.12\%$ & $2.42\%$ & $0.09\%$ & $0.674$ & $6$-$7$ \\ \hline
  \end{tabular}
\label{CNN_DF_com} 
\end{table} 

Table \ref{CNN_DF_com} demonstrates the detection performance of our model in terms of few standard metrics. The positive detection rate and average IOU (Intersection Over Union) values suggest that it is accurate in localizing the person in the image-space. Although it is slower than the MDPM tracker (runs at $6$-$7$ fps), we have found this to be sufficient for following a diver in practice.

\subsection{Results for Human-Robot Communication Framework}\label{sec:DF}
The overall performance of our hand-gesture based human-robot communication framework mostly depends on the accuracy and correctness of the hand gesture recognition module. This is because the FSM-based instruction decoder is deterministic and has a one-to-one gesture-to-instruction mapping. In addition, the robustness of the mapping is ensured by the following transition rules: 
\begin{itemize}
\item State transitions are activated only if the corresponding gesture-tokens are detected for $10$ consecutive frames. Therefore, an incorrect recognition has to happen $10$ consecutive frames to generate an incorrect instruction-token, which is highly unlikely.
\item Also, there are no transition rules (to other states) for incorrect gesture-tokens. Consequently, incorrect instruction-tokens are not going to generate a complete wrong instruction.  
\end{itemize}

In the following sections, we discuss the training processes of different hand gesture recognizers used in our framework and then demonstrate how the interactions happen in practice.

\begin{figure*}[t]
\vspace{1mm}
    \centering
    \begin{subfigure}[t]{0.75\textwidth}
        \includegraphics[width=\linewidth]{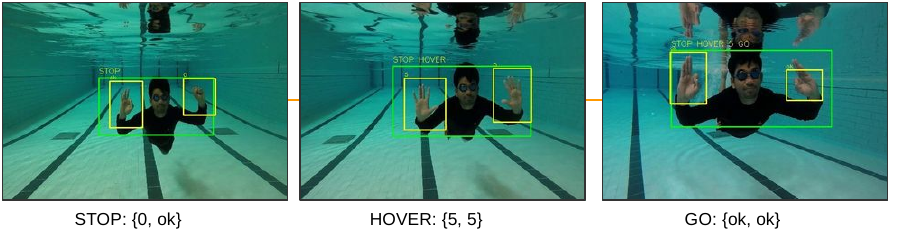}
        \caption{Instructing the robot to stop executing the current program and hover.}
    \end{subfigure}
    
    \vspace{2mm}
    \begin{subfigure}[t]{\textwidth}
        \includegraphics[width=\linewidth]{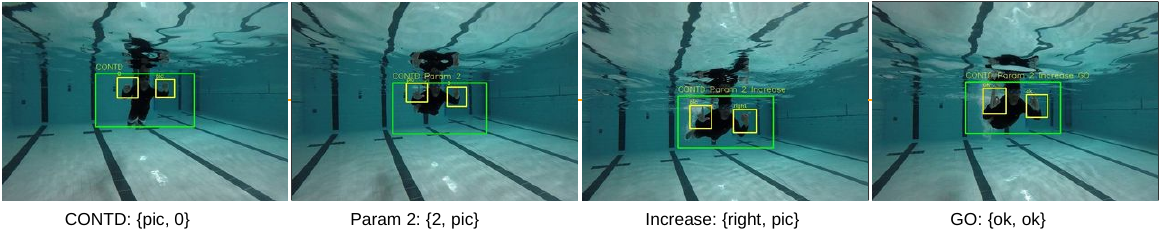}
        \caption{Instructing the robot to continue its current program but increase the value of parameter 2 (by one step).}
    \end{subfigure} 
 \caption{Demonstrations of how the instructions are communicated to the robot using a sequence of hand gestures in our framework. The yellow bounding boxes represent the hand gestures detected by our CNN-based model; here, the green bounding boxes represent the detected region of interest (\textit{i.e.}, the person) in the image. Note that the \{left, right\} hand gestures are ordered as the person's left and right hands.}
 \label{fig:robogest}
\end{figure*}

\subsubsection{Training the Hand Gesture Recognizers} 
As mentioned in Section \ref{HandGest}, we implement three different models for hand gesture recognition; our own CNN-based model with a region selector, Faster RCNN with Inception v2, and SSD with MobileNet v2. 

We presented the architecture of our CNN-based model in Figure \ref{Arch}. Additionally, we illustrated few samples from the training data and associated class (\textit{i.e.}, hand gesture) labels in Figure \ref{data}. The dataset contains over $5$K images per class, and images are annotated to have class labels and bounding boxes. An additional $4$K images are used for validation and a separate $1$K images are used as a test-set. We followed the same training process and an identical setup as presented in Section \ref{sec:deppDF}. It takes about $50$ epochs to train our model with a batch-size of $128$; we refer to Appendix V for visualizing the convergence behavior.

On the other hand, we utilized the pre-trained models for Faster RCNN with Inception v2 and SSD with MobileNet v2 that are provided in the TensorFlow object detection module \cite{tfzoo}. We trained these models with the same dataset and then used them as hand gesture recognizers in our framework. These models are trained for $200$K iterations with the default configurations provided in their APIs.

\begin{table}[ht]
\caption{Performance of our framework on test data using different hand gesture recognizers.}
\centering
\footnotesize
\begin{tabular}{|m{1.95cm}|m{1.4cm}|m{1.1cm}|m{1.35cm}|m{0.7cm}|} \hline
Hand Gesture Recognizer & Total \# of Instructions (Gestures) & Correct Detection & Accuracy (\%) & FPS (robot CPU) \\ \hline \hline
Our Model & $30$ ($162$) & $24$ ($128$) & $80$ ($78$) & $17$-$18$  \\ \hline
Faster RCNN (Inception v2) & $30$ ($162$) & $29$ ($152$) & $96.6$ ($93.8$) & $2$-$3$ \\ \hline
SSD (MobileNet v2) & $30$ ($162$) & $27$ ($144$) & $90$ ($88.8$) & $6$-$7$ \\ \hline 
\end{tabular}
\label{comp}
\end{table}

\subsubsection{Experimental Evaluations}
Figure \ref{fig:robogest} demonstrates how divers can communicate instructions to the robot using a sequence of hand gestures in our framework. As mentioned, the overall success of the operation mostly depends on the correctness of hand gesture recognition. We test our framework extensively using the three different hand gesture recognizers. The test dataset contain a diverse set of $30$ instructions that involves a total of $162$ hand gestures. Table \ref{comp} illustrates the performance of our framework for the different choices of hand gesture recognizers.  

As seen in Table \ref{comp}, our CNN-based model is significantly faster than the state-of-the-art models. However, the detection accuracy is not very good; it correctly detected $24$ out of $30$ instructions with a hand gesture recognition accuracy of $78\%$.    We inspected the failed cases and found the following issues: 
\begin{itemize}
\item In some cases, the diver's hand(s) appeared in front of his face or only partially appeared in the field-of-view. In these cases, not all of the hand(s) appeared in the selected region which eventually caused the gesture recognizer to detect `$ok$'s as `$0$'s, or `$pic$'s as `$1$'s, etc.  

\item Surface reflection and air bubbles often cause problems for the region selector. Although surface reflection is not common in deep water, suspended particles and limited visibility are additional challenges in deep water scenarios.    
\end{itemize}

The state-of-the-art deep visual detectors perform much better in such challenging conditions. As demonstrated in Table \ref{comp}, Faster RCNN correctly detected $29$ out of $30$ instructions with a hand gesture recognition accuracy of $93.8\%$. On the other hand, SSD correctly detected $27$ our of $30$ instructions with an $88.8\%$ hand gesture recognition accuracy. Although these detectors are slower than our model, they are significantly more robust and accurate. 

\begin{figure*}[ht]
\vspace{1mm}
    \centering
    \begin{subfigure}[t]{\textwidth}
        \includegraphics[width=\linewidth]{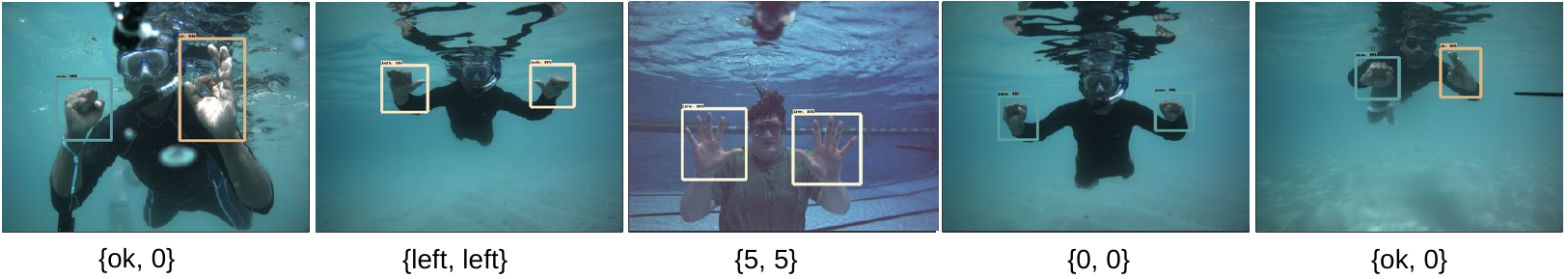}
        \caption{Detections using Faster RCNN (inception v2).}
    \end{subfigure}
    
    \vspace{2mm}
    \begin{subfigure}[t]{\textwidth}
        \includegraphics[width=\linewidth]{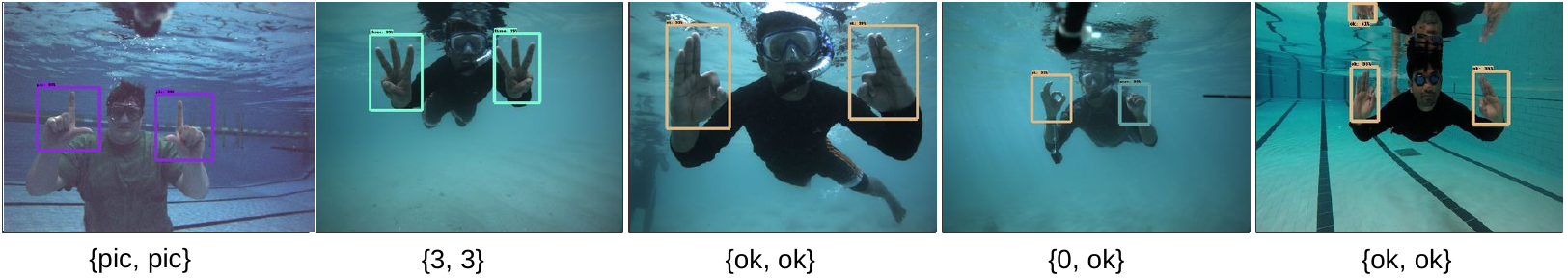}
        \caption{Detections using SSD (MobileNet v2).}
    \end{subfigure} 
   
 \caption{A few snapshots of robust hand gesture recognition by the state-of-the-art object detectors used in our framework.}
 \label{fig:robogestDeep}
\end{figure*}

We have used both Faster RCNN and SSD in our framework for real-time experiments (Figure \ref{fig:robogestDeep}); their slow running times do not affect the overall operation significantly. Detecting hand-gestures is not as time-critical as tracking a diver in real-time; therefore, even $2$-$3$ detections per second is good enough for practical implementations. In the current implementation of our framework, we use SSD (MobileNet v2) as the hand gesture recognizer to balance the trade-offs between performance and running time. 

\subsubsection{Gazebo Simulation}
\begin{figure}[h]
\centering
\includegraphics [width=\linewidth]{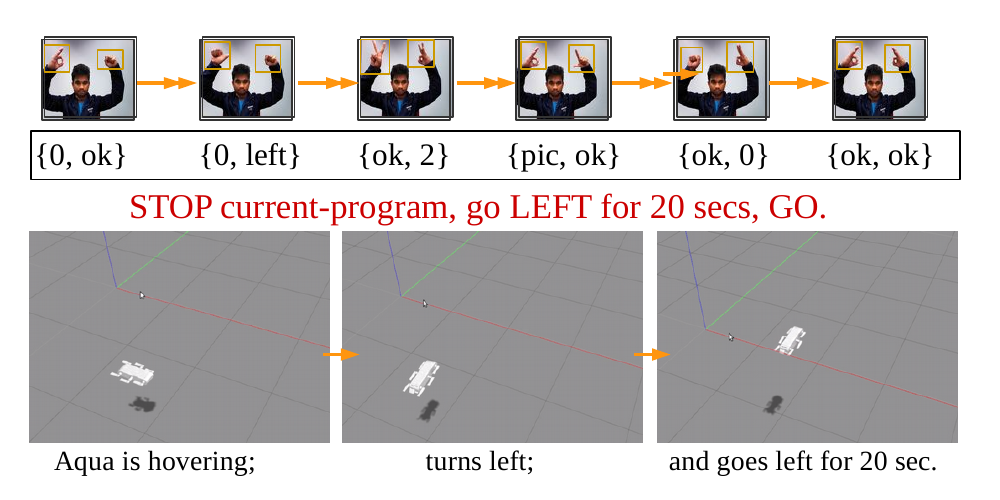}
\vspace{-5mm}
\caption{Controlling an Aqua robot using instructions generated from a sequence of hand gestures performed by a person; the simulation is performed in Gazebo, on the ROS-kinetic platform.}
\label{simu}
\end{figure}

We also performed simulation experiments on controlling an Aqua robot based on the instructions generated from sequences of hand gestures performed by participants. The gesture sequences are captured through a web-cam and the simulation is performed in Gazebo on the ROS Kinetic platform. As illustrated in~Figure \ref{simu}, gesture-tokens are successfully decoded to control the robot. Although a noise-free simulation environment does not pose most challenges that are common in the real world, it does help to set benchmarks for expected performance bounds and is useful in human interaction studies, which is described in the following section.

\subsubsection{Human Interaction Study}
Finally, we performed a human interaction study where the participants are introduced to our hand gesture based framework, the fiducial-based RoboChat framework~\cite{dudek2007visual}, and the RoboChat-Gesture framework \cite{xu2008natural} where a set of discrete motions from a pair of fiducials are interpreted as gesture-tokens. AprilTags~\cite{olson2011apriltag} were used for the RoboChat trials to deliver commands.

A total of ten individuals participated in the study, who were grouped according to their familiarity to robot programming paradigms in the following manner: 
\vspace{1mm}
\begin{itemize}
\item Beginner: participants who are unfamiliar with gesture/fiducial based robot programming ($2$ participants)
\item Medium: participants who are familiar with gesture/fiducial based robot programming ($7$ participants)
\item Expert: participants who are familiar and practicing these frameworks for some time ($1$ participant)
\end{itemize}

This approach is similar to the one used by~\cite{xu2008natural}. In the first set of trials, participants are asked to perform sequences of gestures to generate the following instructions (see Appendix III) in all three interaction paradigms: 
\vspace{1mm}
\begin{enumerate}[$\hspace{4mm}1.$ ]
\item STOP current-program, HOVER for 50 seconds, GO.
\item CONTD current-program, take SNAPSHOTS for 20 seconds, GO.
\item CONTD current-program, Update Parameter 3 to DECREASE, GO.
\item STOP current-program, EXECUTE Program 1, GO.
\end{enumerate}

The second set of trials, participants had to program the robot with complex instructions and were given the following two scenarios:
\begin{enumerate}[\hspace{4mm}a.]
\item \textit{The robot has to stop its current task and execute program 2 while taking snapshots, and}
\item \textit{The robot has to take pictures for 50 seconds and then start following the diver.}
\end{enumerate}

\begin{figure}[ht]
\centering
\includegraphics [width=\linewidth]{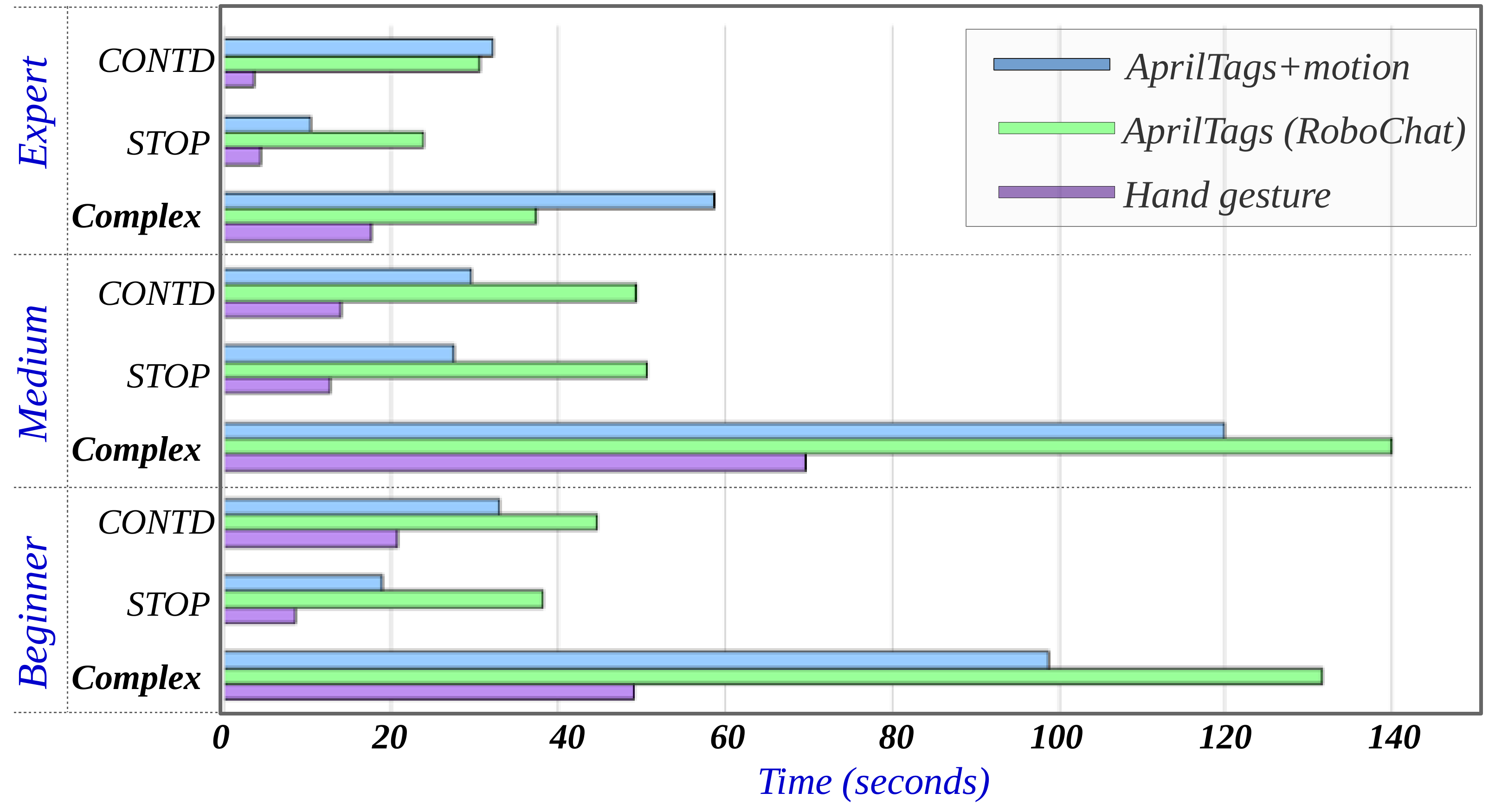}
\vspace{-6mm}
\caption{Comparisons of average time taken to perform gestures to successfully generate different types of programs ($STOP$: instructions $1$ and $4$, $CONTD$: instructions $2$ and $3$, $Complex$: scenarios $a$ and $b$).}
\label{hri}
\vspace{1mm}
\end{figure} 

\vspace{2mm}
For all the experiments mentioned above, participants performed gestures with hands, AprilTags, and discrete motions with AprilTags. Correctness and the amount of time taken were recorded in each case. Figure \ref{hri} shows the comparisons of the average time taken to perform gestures for generating different types of instructions. Participants quickly adopted the hand gestures-to-instruction mapping and took significantly less time to finish programming compared to the other two alternatives. Specifically, participants found it inconvenient and time-consuming to search through all the tags for each instruction token. On the other hand, although performing a set of discrete motions with only two AprilTags saves time, it was less intuitive to the participants. As a result, it still took a long time to formulate the correct gestures for complex instructions, as evident from the results.

One interesting result is that the \textit{beginner} users took less time to complete the instructions compared to \textit{medium} users. This is probably due to the fact that unlike the beginner users, medium users were trying to intuitively interpret and learn the syntax while performing the gestures. However, as illustrated by Table \ref{hri_dudes}, beginner users made more mistakes on an average before completing an instruction successfully. The expert user performed all tasks on the first try, hence only a comparison for beginner and medium users is presented. Since there are no significant differences in the number of mistakes for any types of user, we conclude that simplicity, efficiency, and intuitiveness are the major advantages of our framework over the existing methods.   

\begin{table}[ht]
\caption{Average number of mistakes using [$hand$ $gesture$, $Robochat$, $AprilTtags$ $with$ $motion$] for different users before correctly generating the instruction.}
\centering
\begin{tabular}{|c|c||c|c|} \hline
Instruction & Total \#\ of & Beginner & Medium  \\  
Type & Instructions (Gestures) & User & User  \\ \hline \hline
STOP & $2$ $(10$) & $[2,1,3]$ & $[1,0,1]$ \\ \hline
CONTD & $2$ ($10$) & $[0,0,1]$ & $[0,0,0]$   \\ \hline
Complex & $2$ ($16$) & $[2,3,7]$ & $[2,2,3]$   \\ \hline
\end{tabular}
\label{hri_dudes}
\end{table}

\section{CONCLUSION}
In this paper, we present a number of methodologies for understanding human swimming motion and hand gesture-based instructions for underwater human-robot collaborative applications. At first, we design two efficient algorithms for autonomous diver-following. The first algorithm, named MDPM tracker, uses both spatial- and frequency-domain features to track a diver in the spatio-temporal volume. The second algorithm uses a CNN-based model for robust detection of a diver in the image-space. We also propose a hand gesture-based human-robot communication framework, where a diver can use a set of intuitive and meaningful hand gestures to program new instructions or reconfigure existing program parameters for an accompanying robot on-the-fly. In the proposed framework, CNN-based deep visual detectors provide accurate hand gesture recognition and an FSM-based deterministic model performs robust gesture-to-instruction mapping. 

The accuracy and robustness of the proposed diver-following algorithms and the human-robot interaction framework are evaluated through extensive field experiments. We demonstrate that these modules can be used in real-time for practical applications. In our future work, we plan to explore the feasibilities of using state-of-the-art object detection methods for autonomous diver following. In particular, we aim to investigate real-time diver pose detection to enable a robot to anticipate diver's motion direction as well as their current activity. In addition, we intend to accommodate a larger vocabulary of instructions in our interaction framework and add control-flow operations for more elaborate mission programming.

\addtolength{\textheight}{-0cm}


\bibliography{bibtex/bib/IEEEabrv.bib,bibtex/bib/IEEEexample.bib}{}
\bibliographystyle{IEEEtran}

\newpage
\begin{appendices}
\section{Recursive Formulation of $\mu ^*(T)$}

\begin{equation*}
\footnotesize
\begin{aligned}
\mu ^*(T) &= \argmax_{ w^{(0:T-1)}_{i} }{ P \Big\{ G_{0:T}=w^{(0:T)}_{i} \Big| e_{0:T}  \Big\}} \\
 &= \argmax_{w^{(0:T-1)}_{i}}{ \Big( P \Big\{ e_T \Big| G_{0:T}=w^{(0:T)}_{i} \Big\} } \times \\ 
  & \qquad \qquad \qquad   {P \Big\{ G_{0:T}=w^{(0:T)}_{i} \Big| e_{0:T-1}  \Big\} \Big) } \\
 &=  \argmax_{w^{(0:T-1)}_{i}}{ \Big( P \Big\{ e_T \Big| G_{T}=w^{(T)}_{i} \Big\} \times } \\ 
 & \qquad \qquad \qquad { P\Big\{ G_{T}=w^{(T)}_{i} \Big| G_{T-1}=w^{(T-1)}_{i}  \Big\} \times} \\ 
 & \qquad \qquad \qquad { P \Big\{ G_{0:T-1}=w^{(0:T-1)}_{i} \Big| e_{0:T-1}  \Big\} \Big) }    \\
     &=  P \Big\{ e_T \Big| G_{T}=w^{(T)}_{i} \Big\} \times \\
     & \qquad \qquad \qquad  { \argmax_{w^{(T-1)}_{i}}{ \Big( P\Big\{ G_{T}=w^{(T)}_{i} \Big| G_{T-1}=w^{(T-1)}_{i}  \Big\} }  \times} \\
     & \qquad \qquad \qquad { \argmax_{w^{(0:T-2)}_{i}} {P \Big\{ G_{0:T-1}=w^{(0:T-1)}_{i} \Big| e_{0:T-1}  \Big\} } \Big)} \\
   &= P \Big\{ e_T \Big| G_{T}=w^{(T)}_{i} \Big\} \times \\
   & \qquad \qquad \qquad  { \argmax_{w^{(T-1)}_{i}}{ \Big( P\Big\{ G_{T}=w^{(T)}_{i} \Big| G_{T-1}=w^{(T-1)}_{i}  \Big\} } } \times \\ 
   & \qquad \qquad \qquad \qquad \qquad \qquad  {\mu ^*(T-1)  \Big)  }
\end{aligned}
\label{Recur_old}
\end{equation*} 

\vspace{5mm}
\section{Algorithm for Finding Optimal Motion Direction ($v^*$)}
\begin{algorithm}
\begin{algorithmic}[1]
\State Set values for parameters:  $T$, $M$, $p$, $R$ 
\State Set initial values to dynamic table entries for Markovian \\ $\quad $ transition probabilities \\
$t \leftarrow 0$
\While{Next frame ($f^{(t)}$) is available}
    \State Define windows: $w^{(t)}_{i_t}$ for $i_t = 0:M-1$
    \State Formulate evidence vector: $e_t$    
    \If {$t>0$}
     	\State Update dynamic table entries for: \\ $\qquad \qquad $ $P\Big\{ G_{t} = w^{(t)}_{i_{t}} \Big| G_{t-1} = w^{(t-1)}_{i_{t-1}}  \Big\}$
     	(for all \\ $\qquad \qquad (t, t-1)$  pairs, using Equation \ref{Eq:pair})
    \EndIf
    \State Update dynamic table entries for: $P\Big\{e_{t} \Big| G_{t}=w^{(t)}_{i_{t}}  \Big\}$ \\ $\qquad$ (using Equation \ref{Eq:evidence})
     \If {$t > (T-1)$} 
     	\State Calculate $\mu ^*(T, p)$ using Equation \ref{Recur} 
     	\State Find $\digamma(\mu)$ for each $\mu \in \mu ^*(T, p)$
     	\State Find $v^*$ using Equation \ref{FinalOpti}
     	\State Shift detection window and update $t$
     \EndIf
\EndWhile
\end{algorithmic}
\end{algorithm}

\section{Generating Instructions Using Hand Gestures}
\begin{center}
\includegraphics [width=\linewidth]{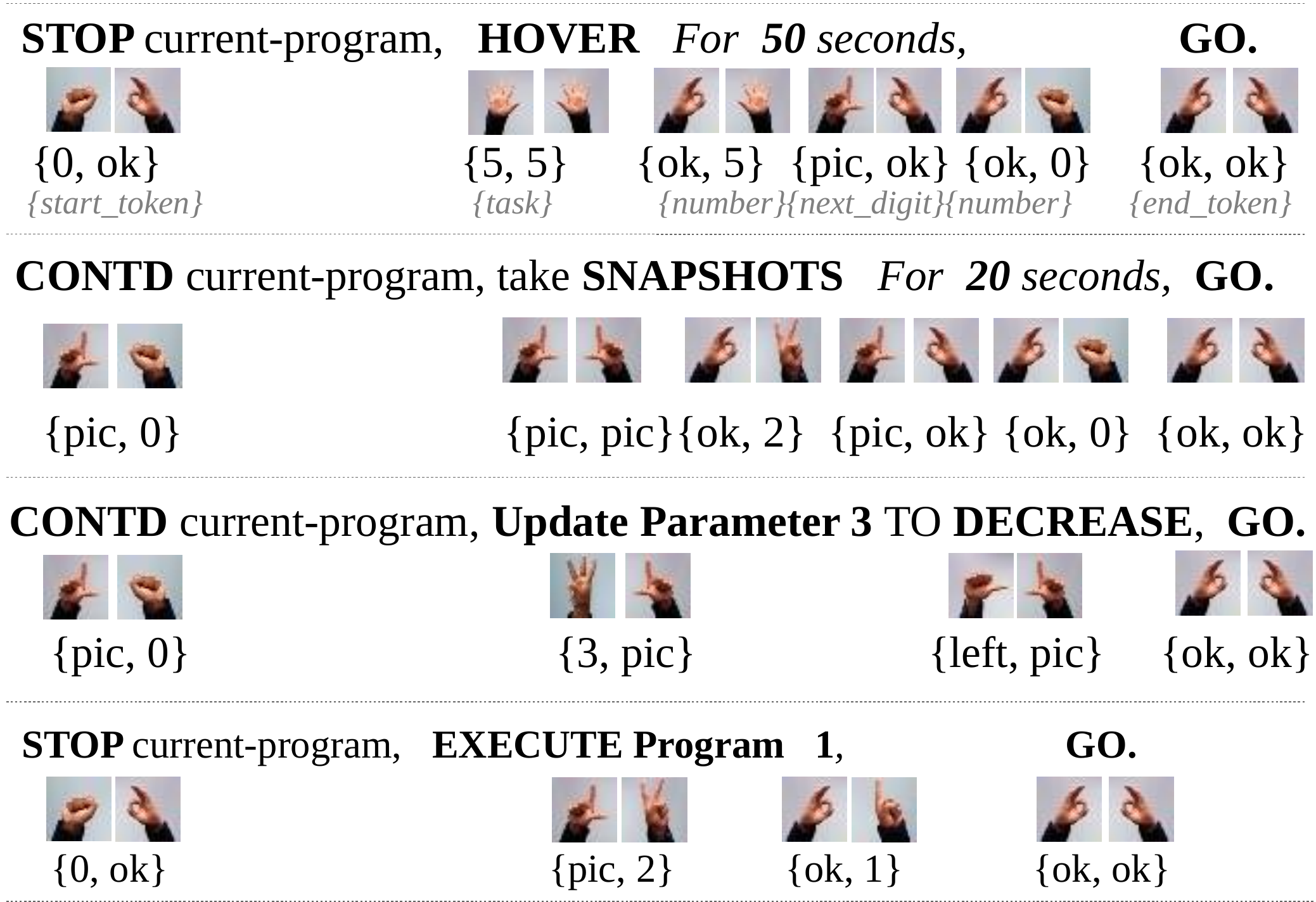}
\end{center}

\vspace{1mm}
\section{Convergence of Our CNN Model for Diver Detection}
\vspace{2mm}
\begin{center}
\includegraphics [width=\linewidth]{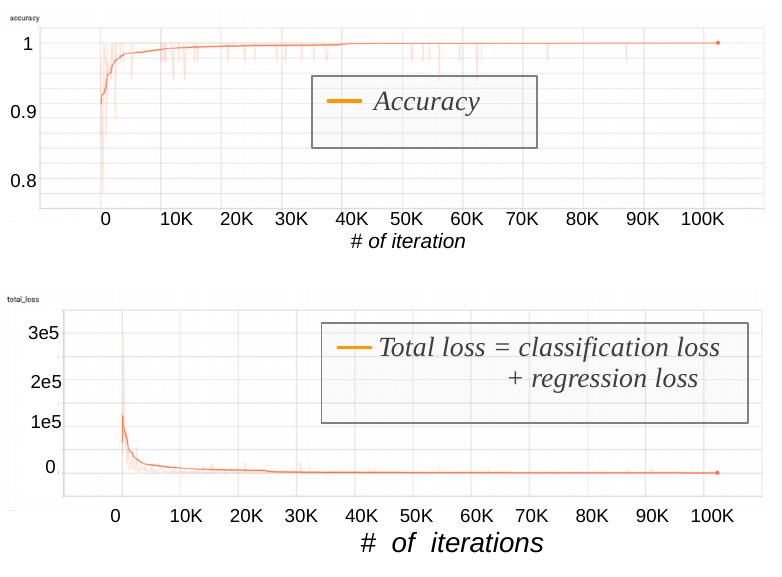}
\end{center}

\vspace{1mm}
\section{Convergence of Our CNN Model for Hand Gesture Recognition}
\vspace{-1mm}
\begin{center}
\includegraphics [width=\linewidth]{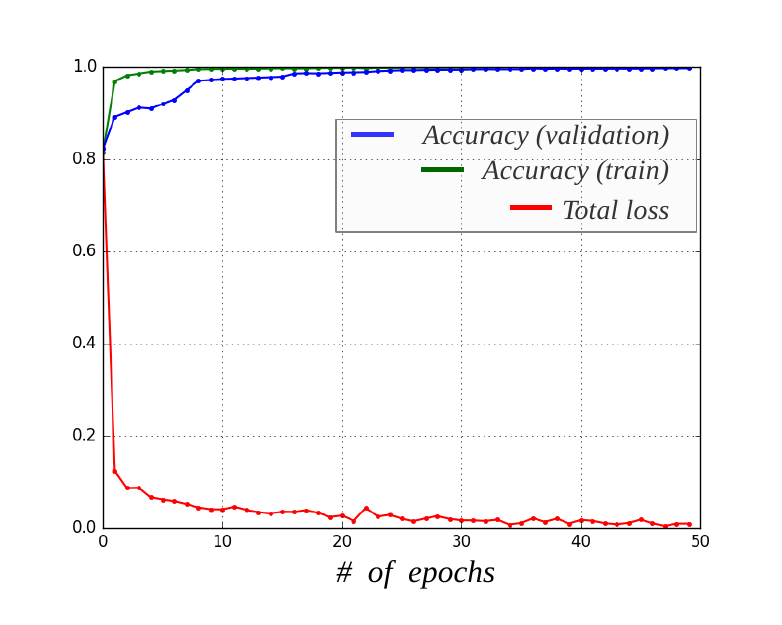}
\end{center}

\end{appendices}

\end{document}